\documentclass[letterpaper]{article} 

\usepackage[hyphens]{url} 
\usepackage{graphicx} 
\usepackage{natbib} 
\usepackage{caption} 
\usepackage{booktabs}
\usepackage{amsmath, amssymb, amsfonts}
\usepackage{multirow}
\usepackage{booktabs}
\usepackage{multirow}
\usepackage{tabularx}
\usepackage{array}
\usepackage{tabularx}
\usepackage{array}

\usepackage[preprint]{aaai2027}

\usepackage{graphicx}
\usepackage{booktabs}
\usepackage{multirow}
\usepackage{amsmath}
\usepackage{amssymb}

\usepackage{booktabs}
\usepackage{threeparttable}
\usepackage{array}
\usepackage{float}
\usepackage{longtable,array,url}

\usepackage{bibunits}

\title{CellPath-Bench: A Multidimensional Benchmark for Whole-Slide \\Cellular Representations in Pathology Foundation Models}

\author {
Bokai Zhao$^\text{1,2,3,\equalcontrib}$, 
Yiyang Zhang$^\text{2,\equalcontrib}$, 
Hanqing Chao$^\text{3,4,\equalcontrib}$, 
Yawei Ma$^\text{2}$, 
Long Bai$^\text{3,4}$, 
Tai Ma$^\text{3,4}$,\\ 
Minfeng Xu$^\text{3,4}$, 
Ming Song$^\text{1,2}$, 
and Tianzi Jiang$^\text{1,2}$\corresponding 
}
\affiliations{
$^\text{1}$School of Artificial Intelligence, University of Chinese Academy of Sciences. \\
$^\text{2}$Brainnetome Center, Institute of Automation, Chinese Academy of Sciences. \\
$^\text{3}$DAMO Academy, Alibaba Group.
$^\text{4}$Hupan Lab.\\
}

\begin{document}

\maketitle

\begin{abstract}
Pathology foundation models (PFMs) are increasingly used as general-purpose backbones, yet existing benchmarks cannot systematically diagnose their whole-slide cellular representation capabilities, including the decodability of cell-type information and the transferability of such information across tissue sections, datasets, and anatomical organs. We introduce \textbf{CellPath-Bench}, a cellular-resolution benchmark that evaluates frozen PFMs themselves. Following quality control of 52 candidate Xenium datasets, we construct a panel of 25 spatially aligned H\&E--Xenium tissue sections spanning 11 organs and 7,079,283 cells, harmonized into fine- and coarse-grained taxonomies. CellPath-Bench samples frozen WSI feature maps at registered nuclear coordinates and evaluates them using standardized multiclass linear probes. Cell Representation Advantage (CRA) measures the within-section advantage of nucleus-anchored representations over patch-level mean pooling, while Cell Representation Transferability (CRT) characterizes the generalization of cell-type decodability across tissue sections, datasets, and organs. We benchmark 30 pathology-specific and general-purpose foundation models through 304,920 runs across spatial readouts, magnifications, taxonomic granularities, and evaluation protocols. The results reveal substantial model-dependent differences in cell-type decodability and its cross-domain generalization, yielding distinct multidimensional capability profiles. CellPath-Bench provides a standardized framework for auditing cellular information in frozen PFM representations.
\end{abstract}

\begin{center}
\textbf{Website:} \url{https://bokai-zhao.github.io/CellPath-benchboard/}
\end{center}

\section{Introduction}

Pathology foundation models (PFMs) learn transferable visual representations through large-scale pretraining on histopathology images and increasingly serve as general-purpose representation backbones in computational pathology~\cite{chnehao2025survey}. As PFMs rapidly diversify in architecture, scale, and pretraining paradigm, systematically characterizing their learned representations has become increasingly important~\cite{PathROB}. Cells are the fundamental units of tissue organization, and differences in cellular identity, composition, and spatial arrangement collectively shape histological phenotypes and tissue microenvironments. For PFMs intended to support diverse pathology applications, preserving cell-discriminative information is therefore an important component of their general-purpose representation capability. Understanding how effectively different PFMs encode such information is essential for characterizing their capabilities and guiding future model development.

\begin{figure}[t]
    \centering
    \includegraphics[
        width=\columnwidth,
        trim=0 0 0 0,
        clip
    ]{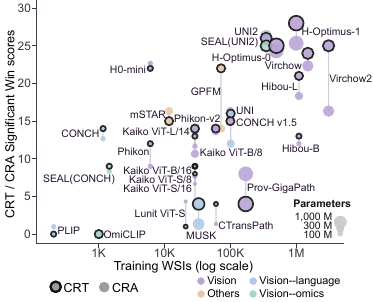}
\caption{
\textbf{CRA--CRT landscape of foundation models.}
Outlined bubbles show CRA significant wins under IS, whereas filled bubbles show CRT, averaged across IOCD, MOCV, and LOOO. The x-axis indicates WSI-equivalent pretraining scale (log scale), and bubble area denotes parameter count. Models without comparable pretraining-scale metadata are omitted.
}
    \label{fig:model_landscape}
\end{figure}

Existing evaluations have primarily established the downstream utility of PFMs through patch-level classification, region-level dense prediction, and whole-slide prediction~\cite{nbme2025benchmarking,chenhao2025pathbench}. These evaluations are indispensable for assessing task-specific applicability, but they do not directly quantify the cell-level representation capability of the frozen PFM itself. Their outcomes jointly depend on the spatial unit of evaluation, feature aggregation, prediction architecture, adaptation strategy, and optimization procedure. A model may perform well on region- or whole-slide-level tasks by exploiting tissue composition, regional texture, or other aggregated evidence, even when local cellular variation is not strongly preserved. Conversely, spatial aggregation may obscure cell-discriminative information that remains available in local features. Higher-level downstream utility and cell-level representation capability are therefore related, but not interchangeable, properties of a PFM. Consequently, downstream performance differences are often difficult to attribute to the frozen representation itself, limiting the diagnostic value of these evaluations for improving PFM architectures and pretraining strategies.

Recent benchmarks have begun to evaluate PFMs at finer cellular and spatial scales from complementary perspectives. PFM-DenseBench~\cite{PFM-DenseBench} evaluates how token-level representations support dense prediction under different adaptation strategies; PathoCellBench~\cite{pathocellbench} assesses cellular phenotyping and cross-domain transferability within localized image fields; and SpaPath-Bench~\cite{SpaPath-Bench} examines regional spatial organization using representations aligned with spatial-transcriptomic spots. Together, these studies extend PFM evaluation beyond conventional patch- and whole-slide-level tasks. However, a systematic comparison of the cell-level representation capability of diverse frozen foundation models under a common, multi-organ cellular reference remains lacking. In particular, it remains insufficiently characterized how much cell-type information is linearly decodable from nucleus-anchored representations and how reliably this capability transfers across tissue sections, datasets, and anatomical organs.

To address these two limitations, we introduce \textbf{CellPath-Bench}, a multidimensional benchmark that treats frozen PFMs themselves as the objects of measurement. First, to separate frozen representation quality from task-specific adaptation, CellPath-Bench keeps all encoders frozen, samples their spatial feature fields at registered nuclear coordinates, and evaluates the resulting representations using a unified multiclass linear probe. Second, to support systematic evaluation under a common, multi-organ cellular reference, we construct a quality-controlled H\&E--Xenium panel with harmonized cell-type taxonomies and define evaluation protocols across tissue sections, datasets, and anatomical organs. Starting from 52 candidate Xenium~\cite{Xenium} datasets, we perform systematic quality control and construct an evaluation panel of 25 spatially registered tissue sections spanning 11 organs and 7,079,283 cells. These cells are harmonized into fine- and coarse-grained cell-type taxonomies, providing a standardized, molecularly informed cellular reference. The cellular reference, spatial readout definitions, and classifier capacity are held constant across heterogeneous model architectures. Under this controlled setting, Nuc Macro-F1 measures how much cell-type information is linearly decodable from the nucleus-anchored representation of each frozen PFM. Within individual tissue sections, Cell Representation Advantage (CRA) complements this absolute performance by measuring the Macro-F1 difference between nucleus-anchored representations and patch-level mean pooling under intra-section spatial transfer (IS). Beyond individual sections, intra-organ cross-dataset transfer (IOCD), multi-organ cross-validation (MOCV), and leave-one-organ-out transfer (LOOO) evaluate how reliably nucleus-anchored cell-type information generalizes across tissue sections, datasets, and anatomical organs. Cell Representation Transferability (CRT) provides a panel-relative summary of model performance across these cross-domain protocols.

Our large-scale evaluation of 30 pathology-specific and general-purpose foundation models reveals substantial variation in cell-type decodability and cross-domain transferability. As summarized by the joint CRA--CRT landscape in Fig.~\ref{fig:model_landscape}, within-section Cell Representation Advantage and cross-domain Cell Representation Transferability are not fully aligned, revealing complementary properties of frozen PFM representations. Positive CRA across all 30 models indicates that nucleus-anchored sampling retains cell-type information that is less accessible after patch-level spatial averaging. Beyond these two core measures, magnification, taxonomic granularity, and contextual feature integration further differentiate model behavior, together revealing multidimensional representation profiles that cannot be captured by a single aggregate ranking. By separating absolute cell-type decodability, sensitivity to spatial readout, and cross-domain generalization, CellPath-Bench provides a more detailed diagnosis of frozen PFM representations than a single downstream task score.

Our main contributions are:
\begin{itemize}

    \item \textbf{A coordinate-aligned framework for measuring cell representations in frozen foundation models.} We sample frozen WSI feature fields at registered nuclear coordinates and evaluate different spatial readouts using a unified linear probe. In this framework, Nuc performance measures absolute cell-type decodability, CRA measures sensitivity to spatial readout within tissue sections, and CRT summarizes relative model performance across tissue sections, datasets, and anatomical organs.

    \item \textbf{A quality-controlled, multi-organ cellular reference.} From 52 candidate Xenium datasets, we construct an evaluation panel of 25 spatially registered H\&E--Xenium tissue sections spanning 11 organs and 7,079,283 cells, with molecularly informed annotations, harmonized fine- and coarse-grained taxonomies, and independent histological concordance assessment.

    \item \textbf{A large-scale, multidimensional evaluation of foundation models.} We benchmark 30 pathology-specific and general-purpose foundation models through 304,920 controlled linear-probe runs across spatial readouts, magnifications, taxonomic granularities, and evaluation protocols. This evaluation systematically assesses how much cell-type information is linearly decodable from each model within tissue sections and how reliably this information transfers across tissue sections, datasets, and anatomical organs.
\end{itemize}

\begin{figure*}[!t]
    \centering
    \includegraphics[
        width=\textwidth
    ]{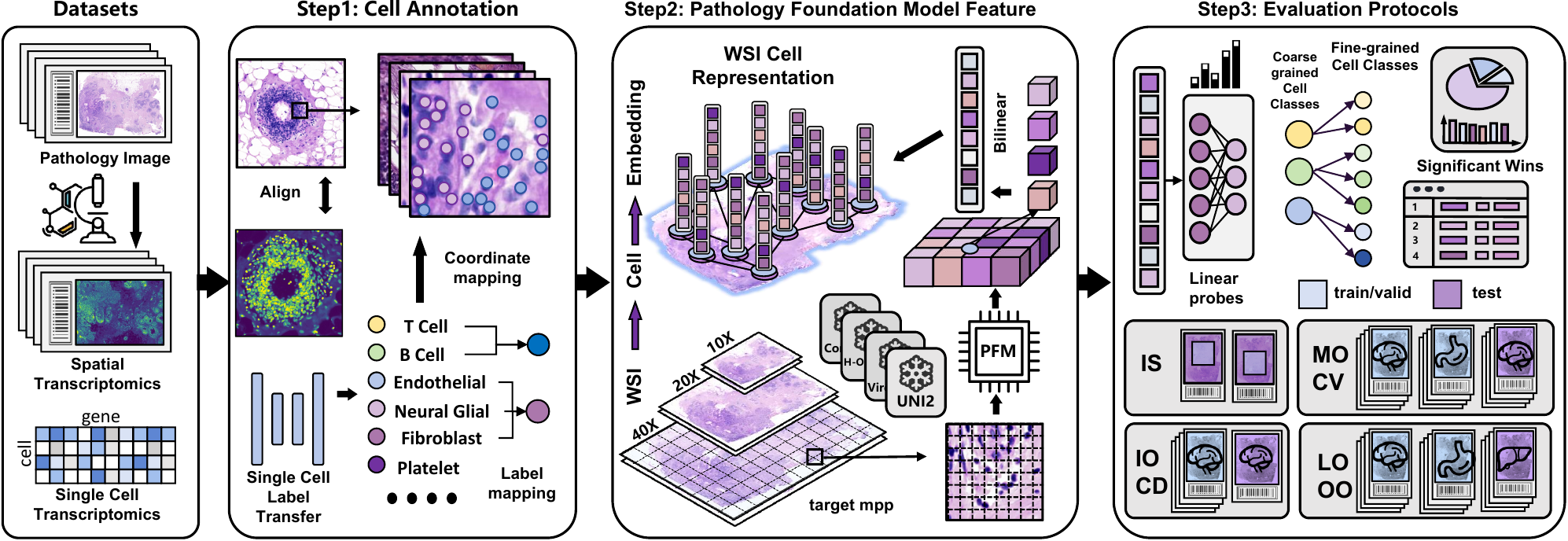}
    \caption{
    Overview of CellPath-Bench, a benchmark for auditing cellular representations in pathology foundation models.
    }
    \label{fig:cellpath_overview}
\end{figure*}

\section{CellPath-Bench}

\subsection{Benchmark Overview}

As illustrated in Fig.~\ref{fig:cellpath_overview}, CellPath-Bench separates the evaluation workflow into three components: a registered cellular reference, a coordinate-aligned representation interface, and protocol-specific linear probing. Each tissue section provides an H\&E image, registered nuclear coordinates, and harmonized Xenium-derived cell labels. For each PFM and magnification, image patches and their nuclear coordinates are mapped to the model's spatial feature grid, from which fixed readout operators construct \emph{Nuc}, \emph{Mean}, and \emph{Cat} representations. These representations are evaluated using a common multiclass linear probe under IS, IOCD, MOCV, and LOOO, with data partitions, readout definitions, probe configurations, and evaluation metrics held constant across models. The following subsections specify each component in detail.

\subsection{Benchmark Data and Task Construction}

\paragraph{Source data and single-cell references.}
We collected public Xenium datasets from 10x Genomics with high-resolution H\&E images spatially registered to the corresponding transcriptomic measurements. Each dataset represents one tissue section and provides the registered H\&E image, cell coordinates, and cell-by-gene expression matrix. For annotation, we assembled 14 labeled scRNA-seq references: one normal-tissue atlas and 13 cancer-specific references. Same-patient or tissue-matched references were prioritized; otherwise, references were matched by organ and disease state. Table~\ref{tab:scrna_reference_panel} summarizes the reference panel. Original Xenium dataset names, source URLs, and section-to-reference assignments are provided in the Supplementary Material.

\begin{table}[h]
\centering
\caption{
scRNA-seq references used for annotation.
}
\label{tab:scrna_reference_panel}
\scriptsize
\setlength{\tabcolsep}{3pt}
\begin{tabular}{@{}p{0.26\columnwidth}p{0.39\columnwidth}rr@{}}
\toprule
Tissue / disease state & scRNA-seq reference & Ref. cells & Sections \\
\midrule

{Breast cancer}
& {\cite{wu2021single}}
& {100064}
& {19} \\
{Normal tissue}
& {\cite{tabulasapiens376thetabulasapiens}}
& {198633}
& {8} \\
{Ovarian cancer}
& {\cite{vazquez2022ovarian}}
& {927205}
& {4} \\
{Lung cancer}
& {\cite{kim2020single}}
& {208506}
& {3} \\
{Lymphoid hyperplasia}
& {\cite{tabulasapiens376thetabulasapiens}}
& {327472}
& {3} \\
{Melanoma}
& {\cite{tirosh2016dissecting}}
& {4645}
& {3} \\
{Pancreatic cancer}
& {\cite{moncada2020integrating}}
& {3659}
& {3} \\
{Colorectal cancer}
& {\cite{pelka2021spatially}}
& {370115}
& {2} \\
{Kidney cancer}
& {\cite{zhang2021single}}
& {32352}
& {2} \\
{Cervical cancer}
& {\cite{li2021single}}
& {25642}
& {1} \\
{Prostate cancer}
& {\cite{hirz2023dissecting}}
& {156987}
& {1} \\
{Glioblastoma}
& {\cite{ruiz2025charting}}
& {338564}
& {1} \\
{Leukemia}
& {\cite{caron2020single}}
& {37320}
& {1} \\
{Liver cancer}
& {\cite{ma2019tumor}}
& {9946}
& {1} \\
\bottomrule
\end{tabular}
\end{table}

\paragraph{Reference-guided molecular annotation.}
We harmonized reference-specific labels into 68 molecular cell types using a marker knowledge base curated from more than 300 publications and covering primary, secondary, and negative markers. These labels were transferred from the matched scRNA-seq references to the Xenium sections using SpCAST~\cite{zhang2026spcast}. Each section was manually reviewed for expected marker patterns and relative marker enrichment, without applying a universal numerical threshold; all 52 candidates formed the annotation-qualified pool. For histology-based evaluation, the 68 molecular cell types were deterministically mapped to the nine-class \(T_{\mathrm{fine}}\) and three-class \(T_{\mathrm{coarse}}\), defining the two classification tasks. Complete taxonomy mappings and cell-eligibility rules are provided in the Supplementary Material.

\paragraph{Quality assessment and benchmark panel construction.}
After excluding sections with fewer than two valid classes, we performed a two-stage quality assessment. First, we inspected the provider-supplied spatial alignment by overlaying Xenium cell coordinates on 20 sampled H\&E regions per section; no image re-registration was performed. Second, CellViT++~\cite{horst2026cellvit++} independently predicted nuclear morphology from H\&E. Its predictions and the Xenium-derived labels were matched by mutual nearest neighbors and mapped to a shared three-class space, with section-level balanced accuracy used as the morphology--annotation concordance score. This score was used solely to rank sections: CellViT++ neither generated nor modified the Xenium-derived labels, and its detection or matching results did not affect cell-level eligibility. The 25 highest-scoring sections formed the primary panel, \(S_{25}\), spanning 11 organs and 7,079,283 spatially indexed cells. Unless otherwise stated, all analyses use \(S_{25}\). Robustness to panel size was evaluated using the nested higher-concordance subsets \(S_{10}\subset S_{15}\subset S_{20}\subset S_{25}\). CellViT++ configuration, three-class mapping, cell matching, section-level scores, panel composition, and robustness results are detailed in the Supplementary Material.

\subsection{WSI-to-Cell Representation Framework}
\label{sec:wsi_to_cell_framework}

\subsubsection{Problem formulation.}
Let \(I_s^r\) denote the H\&E image of tissue section \(s\) at magnification \(r\). Under cell-type taxonomy \(t\in\{T_{\mathrm{fine}},T_{\mathrm{coarse}}\}\), each evaluable cell \(i\) has a registered nuclear coordinate \(\mathbf p_{si}\) and a task-specific label \(y_{si}^{(t)}\). Given a frozen encoder \(E_m\) and a readout operator \(R_q\), we extract the cell representation
\begin{equation}
\mathbf z_{si}^{m,r,q}
=
R_q\!\left(E_m(I_s^r),\mathbf p_{si}\right).
\label{eq:wsi_to_cell_readout}
\end{equation}

We denote an evaluation condition by \(\omega=(r,q,t,\pi)\), where \(\pi\) specifies the evaluation protocol. Each protocol defines a basic evaluation unit \(b\) together with its training, validation, and test partitions. For each \((m,\omega,b)\), a new multiclass linear probe is fitted on the training representations, with model selection performed on the validation partition, and evaluated once on the corresponding test partition. The encoder remains frozen throughout the entire procedure. Under this standardized probing procedure, performance operationalizes the amount of cell-type information that is linearly decodable from the frozen foundation-model representation, without task-specific adaptation of the encoder.

\subsubsection{Coordinate-aligned cellular feature extraction.}

At magnification $r$, each WSI $I_s^r$ is partitioned into non-overlapping patches matching the native input size of encoder $m$, with boundary patches padded as needed. Each cell is assigned to the patch containing its registered nuclear center. If the center lies exactly on a patch boundary, the adjacent patch containing the largest portion of the nuclear area is selected. Let $j=j_{m,r}(s,i)$ denote the patch assigned to cell $i$ in tissue section $s$.

For each patch $\mathbf{X}_{sj}^{m,r}$, the frozen encoder produces a spatial token grid $\mathbf{F}_{sj}^{m,r}\in\mathbb{R}^{H_m\times W_m\times D_m}$ and, when supported by the architecture, a class token $\mathbf{c}_{sj}^{m,r}\in\mathbb{R}^{D_m}$. The registered nuclear coordinate $\mathbf{p}_{si}$ is mapped from the WSI coordinate system to a continuous location $\mathbf{u}_{si}^{m,r}$ on the spatial token grid of the assigned patch. We then derive three base readouts:
\begin{equation}
\begin{aligned}
\mathbf{z}_{si}^{m,r,\mathrm{Nuc}} &= \operatorname{Bilinear}\left(\mathbf{F}_{sj}^{m,r},\mathbf{u}_{si}^{m,r}\right),\\
\mathbf{z}_{si}^{m,r,\mathrm{Mean}} &= \operatorname{MeanPool}\left(\mathbf{F}_{sj}^{m,r}\right),\\
\mathbf{z}_{si}^{m,r,\mathrm{Cls}} &= \mathbf{c}_{sj}^{m,r}.
\end{aligned}
\label{eq:base_readouts}
\end{equation}
The $\mathrm{Nuc}$ readout samples the feature grid at the registered nuclear center and therefore uses the coordinate as a spatial anchor; it does not imply a nucleus-restricted receptive field. In contrast, $\mathrm{Mean}$ and $\mathrm{Cls}$ provide patch-level summaries through spatial mean pooling and the class token, respectively. For encoders without a native class token, the $\mathrm{Cls}$ readout and its derived fusions are omitted.

To assess whether patch-level context complements the nucleus-anchored representation, we combine $\mathrm{Nuc}$ with each available global readout $g\in\{\mathrm{Mean},\mathrm{Cls}\}$ using either element-wise addition, $\mathrm{Nuc}+g$, or feature concatenation, $\mathrm{Nuc}\Vert g$. The two components are combined directly without normalization, rescaling, or projection. This yields up to seven representation modes: $\mathrm{Nuc}$, $\mathrm{Mean}$, $\mathrm{Cls}$, $\mathrm{Nuc}+\mathrm{Mean}$, $\mathrm{Nuc}+\mathrm{Cls}$, $\mathrm{Nuc}\Vert\mathrm{Mean}$, and $\mathrm{Nuc}\Vert\mathrm{Cls}$.

\subsection{Evaluation Protocols}
\label{sec:evaluation_protocols}
\paragraph{Hierarchical transfer protocols.}
While individual cells serve as the fundamental classification samples, we define our test partitions at ascending levels of spatial and anatomical hierarchy: spatial regions, complete tissue sections, and complete organs. To rigorously prevent data leakage, cells sharing a held-out spatial or anatomical unit are strictly excluded from probe fitting. We characterize model transferability across these hierarchies using four distinct evaluation protocols:

\begin{itemize}
    \item \textbf{IS (Intra-section spatial transfer):} Evaluates generalization across spatially separated regions within a single tissue section. For a given section and spatial split, cells inside a predefined region of interest (ROI) are used for training and validation, whereas cells outside this region form the test set. The test score is computed from all valid cells outside the training region.
    
    \item \textbf{IOCD (Intra-organ cross-dataset transfer):} Assesses cross-section generalization within a specific, known organ. Under this protocol, one complete tissue section from an eligible organ is held out entirely for testing, while the remaining sections of that same organ supply the training and validation data. Crucially, no cell or image patch from the test section is exposed during probe fitting or model selection.
    
    \item \textbf{MOCV (Multi-organ cross-validation):} Evaluates transferability to held-out tissue sections under heterogeneous, multi-organ training. Complete tissue sections are assigned to a fixed, organ-aware cross-validation manifest, keeping all cells from a given section in the same fold. Probes are fitted on the remaining folds, and a single score is computed by pooling test predictions from all sections in the held-out fold. This evaluates multi-organ transfer without requiring the test organs themselves to be entirely unseen.
    
    \item \textbf{LOOO (Leave-one-organ-out transfer):} Measures zero-shot cross-organ transfer. For each target organ, all of its corresponding tissue sections are completely excluded from training and validation. All valid cells from the held-out organ are pooled and scored together, without averaging section-level metrics within the organ.
\end{itemize}

\paragraph{Metrics, aggregation, and statistical analysis.}
Each evaluation unit produces one score from its complete test partition. Macro-F1 is the primary metric, while Macro-AUROC is recorded as a secondary metric.
Protocol summaries balance the anatomical or cross-validation units rather than pooling all cells or runs. For IS, split--seed scores are first averaged within each tissue section and then across tissue sections of the same organ. IOCD averages seeds and held-out tissue sections within each organ. MOCV averages seeds within each test fold, and LOOO averages seeds within each held-out organ. Let $A_{m,u}^{\pi}$ denote the resulting organ-level score for IS, IOCD, and LOOO, or fold-level score for MOCV. We report as $\overline{x}_{m,\omega}\pm\sigma_{m,\omega}$. Thus, the reported standard deviation reflects variation across organs or folds rather than across pooled cells.
\begin{equation}
\overline{x}_{m,\omega} = \operatorname{Avg}_{u\in\mathcal{U}_\pi} A_{m,u}^{\pi}, \quad \sigma_{m,\omega} = \operatorname{SD}_{u\in\mathcal{U}_\pi} A_{m,u}^{\pi}.
\label{eq:protocol_summary}
\end{equation}

Statistical comparisons are conducted separately for each evaluation condition and for the metric underlying the corresponding analysis. For each model pair, we apply a two-sided paired Wilcoxon signed-rank test over their common valid evaluation units. A model records a significant win over another model when the $p$-value is below $0.05$ and the median paired difference favors that model. On large-scale benchmarks, significant-win counts provide a more stable assessment for model ranking than the protocol-level mean and allow for straightforward integration across different evaluation protocols.

\begin{table}[t]
\centering
\caption{
  The 30 foundation models evaluated in CellPath-Bench.
}
\label{tab:evaluated_models}
\scriptsize
\setlength{\tabcolsep}{2.5pt}
\renewcommand{\arraystretch}{0.92}

\begin{tabularx}{\columnwidth}{
    @{}
    l
    >{\raggedright\arraybackslash}X
    >{\raggedleft\arraybackslash}p{0.12\columnwidth}
    >{\raggedleft\arraybackslash}p{0.18\columnwidth}
    @{}
}
\toprule
Type
& Model
& Params.
& \shortstack[r]{Pre. WSI} \\
\midrule

\multirow[t]{20}{*}{PV}
& H-Optimus-1~\cite{hoptimus1} & 1100M & 1000K \\
& H-Optimus-0~\cite{hoptimus0}  & 1100M & 500K \\
& Prov-GigaPath~\cite{Prov-GigaPath}       & 1100M & 171K \\
& Virchow2~\cite{Virchow2}    & 632M  & 3100K \\
& Virchow~\cite{Virchow}       & 632M  & 1500K \\
& UNI2~\cite{UNI}                  & 632M  & 350K \\
& GPFM~\cite{GPFM}                    & 307M  & 72K \\
\bottomrule
\end{tabularx}
\end{table}

\section{Experiments and Results}
\label{sec:experiments_results}

\subsection{Experimental Setup}
\label{sec:experimental_setup}

\subsubsection{Evaluated foundation models.}
\label{sec:evaluated_models}
We evaluated 30 foundation models spanning four pretraining paradigms (Table~\ref{tab:evaluated_models}): pathology vision (PV), pathology vision--language (PVL), pathology vision--omics (PVO), and general-purpose vision (GV). The panel covers diverse architectures, parameter scales, and supervision sources, enabling a systematic comparison of representations learned from pathology images, paired biomedical data, and general-domain visual corpora. All encoders were kept frozen and evaluated using a unified linear-probing procedure. To facilitate scale comparisons, Table~\ref{tab:evaluated_models} reports approximate WSI-equivalent (WSI-eq.) pretraining counts; for models reporting only patch or image--text pair counts, we apply a standardized conversion ratio of 1,000 patches/pairs per WSI. Detailed model architectures, pretraining corpora, checkpoint sources, and encoder-specific preprocessing are provided in the Supplementary Material.

\subsubsection{Experimental scope.}
\label{sec:experimental_scope}
\begin{table}[t]
\centering
\caption{
  Derivation of evaluation units, random seeds, and total linear-probe runs across the four protocols. The constant multipliers in the run calculations correspond to the 30 models, 3 magnifications, 2 taxonomies, and 7 representation modes evaluated.
}
\label{tab:experimental_scope}
\scriptsize
\renewcommand{\arraystretch}{0.95}

\begin{tabular*}{\columnwidth}{
    @{\extracolsep{\fill}}
    l
    l
    c
    l
    @{}
}
\toprule
Protocol
& Evaluation units
& Seeds
& Runs \\
\midrule

IS
& 25 sections
& 5
& $30{\times}3{\times}2{\times}25{\times}5{\times}7
   =157{,}500$ \\

IOCD
& 23 folds (9 organs)
& 3
& $30{\times}3{\times}2{\times}23{\times}3{\times}7
   =86{,}940$ \\

MOCV
& 5 folds
& 3
& $30{\times}3{\times}2{\times}5{\times}3{\times}7
   =18{,}900$ \\

LOOO
& 11 folds (11 organs)
& 3
& $30{\times}3{\times}2{\times}11{\times}3{\times}7
   =41{,}580$ \\

\midrule
\multicolumn{4}{r}{
    Total:
    $157{,}500+86{,}940+18{,}900+41{,}580
    =\mathbf{304{,}920}$
} \\
\bottomrule
\end{tabular*}
\end{table}
For each protocol-specific evaluation unit, we trained an independent multiclass linear probe while keeping the foundation-model encoder frozen. We define one run by a fixed combination of foundation model, magnification, representation mode, cell-type taxonomy, evaluation protocol, evaluation unit, and random seed. Random seeds were involved in data partitioning rather than serving solely as repeated probe initializations. All probes followed the same training procedure; optimization details and computational configurations are provided in the Supplementary Material.

The evaluation covered 30 foundation models at three magnifications ($40\times$, $20\times$, and $10\times$) under both the fine-grained ($T_{\mathrm{fine}}$) and coarse-grained ($T_{\mathrm{coarse}}$) taxonomies, testing the seven representation modes defined in Section~\ref{sec:wsi_to_cell_framework}. Because the spatial and anatomical definitions vary by protocol, the exact number of evaluation units and data-split seeds differs accordingly. Table~\ref{tab:experimental_scope} details these protocol-specific configurations and the resulting run-count derivation, yielding a total of $304{,}920$ independent linear-probe runs.

\subsection{Results}
\label{sec:results}

\subsubsection{Multidimensional Analysis of Cell-Level Representations}
\label{sec:multidimensional_analysis}

\begin{figure*}[t]
    \centering
    \includegraphics[width=\textwidth]{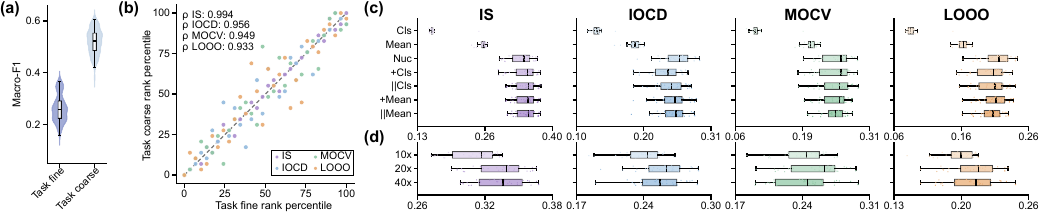}
\caption{
\textbf{Effects of taxonomy, spatial readout, and magnification on cell-level representations.}
\textbf{(a)} Macro-F1 under the fine- and coarse-grained taxonomies using \emph{Nuc} at $20\times$.
\textbf{(b)} Corresponding model-rank concordance; the dashed line indicates identical ranks.
\textbf{(c)} Fine-grained Macro-F1 across seven spatial readouts at $20\times$.
\textbf{(d)} Fine-grained Macro-F1 across three magnifications using \emph{Nuc}.
Each point represents one model after protocol-specific balanced aggregation.
}
    \label{fig:measurement_calibration}
\end{figure*}

We examined how taxonomic granularity, spatial readout, and image magnification affect cell-level representation evaluation (Fig.~\ref{fig:measurement_calibration}).
The coarse-grained taxonomy consistently yielded higher Macro-F1 than the fine-grained taxonomy. Nevertheless, model rankings remained highly consistent, with Spearman's $\rho$ of $0.994$, $0.956$, $0.949$, and $0.933$ for IS, IOCD, MOCV, and LOOO, respectively. We therefore retain both taxonomies to evaluate complementary levels of cellular detail.
At $20\times$, \emph{Nuc} consistently outperformed \emph{Mean} across both taxonomies and all four protocols. The \emph{Cls} readout was less informative, while combining \emph{Nuc} with \emph{Mean} or \emph{Cls} provided no consistent improvement. We therefore use \emph{Nuc} as the primary cell-level readout and \emph{Mean} as its patch-level reference.
Across protocols, performance at $20\times$ remained close to that at $40\times$, whereas $10\times$ showed a clearer decline. Under IS, performance retention relative to $40\times$ was $99.6\%$ at $20\times$ and $92.9\%$ at $10\times$. These results indicate that cell-type information is largely preserved at $20\times$ but more strongly attenuated at $10\times$.
Together, these results establish the evaluation settings for the following analyses. We next quantify the within-section advantage of nucleus-anchored representations over patch-level averaging.

\begin{table}[t]
\centering
\caption{
\textbf{Within-section Cell Representation Advantage at $20\times$.}
\emph{Nuc} and \emph{Mean} report organ-balanced Macro-F1 (mean $\pm$ SD across 11 organs), and CRA follows Eq.~\ref{eq:cra}. Sig.\ wins are based on paired Wilcoxon signed-rank tests ($p<0.05$). Values are percentage points; best and second-best results are \textbf{bolded} and \underline{underlined}.
}
\label{tab:is_20x_cellgain}
\resizebox{\columnwidth}{!}{
\begin{tabular}{lcccccc}
\toprule


\multirow{2}{*}{Model}
& \multicolumn{2}{c}{$T_{\mathrm{fine}}$ (9)}
& \multicolumn{2}{c}{$T_{\mathrm{coarse}}$ (3)}
& \multirow{2}{*}{CRA$\uparrow$}
& \multirow{2}{*}{\shortstack{Sig.\\wins$\uparrow$}} \\
\cmidrule(lr){2-3} \cmidrule(lr){4-5}
& \emph{Nuc} & \emph{Mean}
& \emph{Nuc} & \emph{Mean}
& & \\
\midrule

H-Optimus-1 & \textbf{36.73{\scriptsize$\pm$8.51}} & \textbf{26.95{\scriptsize$\pm$6.52}} & \textbf{56.88{\scriptsize$\pm$8.96}} & \textbf{47.81{\scriptsize$\pm$5.78}} & \textbf{10.14{\scriptsize$\pm$6.05}} & \textbf{29} \\
Virchow2 & 36.21{\scriptsize$\pm$8.25} & 26.45{\scriptsize$\pm$5.94} & 55.90{\scriptsize$\pm$8.86} & 45.97{\scriptsize$\pm$5.82} & \underline{10.12{\scriptsize$\pm$5.82}} & \underline{25} \\
H-Optimus-0 & 36.38{\scriptsize$\pm$8.35} & \underline{26.80{\scriptsize$\pm$6.23}} & \underline{56.66{\scriptsize$\pm$8.84}} & \underline{47.68{\scriptsize$\pm$5.38}} & 9.96{\scriptsize$\pm$5.89} & 25 \\
UNI2 & \underline{36.41{\scriptsize$\pm$8.60}} & 26.60{\scriptsize$\pm$5.83} & 56.40{\scriptsize$\pm$8.94} & 47.42{\scriptsize$\pm$6.00} & 9.94{\scriptsize$\pm$6.03} & 25 \\
SEAL(UNI2) & 36.41{\scriptsize$\pm$8.59} & 26.72{\scriptsize$\pm$6.03} & 56.40{\scriptsize$\pm$8.94} & 47.44{\scriptsize$\pm$5.97} & 9.89{\scriptsize$\pm$6.03} & 25 \\
Virchow & 35.91{\scriptsize$\pm$8.39} & 26.64{\scriptsize$\pm$6.33} & 56.07{\scriptsize$\pm$8.87} & 46.96{\scriptsize$\pm$5.62} & 9.82{\scriptsize$\pm$5.75} & 24 \\
\bottomrule
\end{tabular}
}
\end{table}

\begin{table*}[t]
\centering
\caption{
\textbf{Cross-domain evaluation of Cell Representation Transferability at $20\times$.}
Scores are Macro-F1 and Macro-AUROC (mean $\pm$ SD) using \emph{Nuc}. Sig.\ wins are computed under $T_{\mathrm{fine}}$ using paired Wilcoxon signed-rank tests ($p<0.05$), and CRT follows Eq.~\ref{eq:crt}. Best and second-best results are \textbf{bolded} and \underline{underlined}.
}
\label{tab:cross}
\resizebox{\textwidth}{!}{
\begin{tabular}{lccccccccccccccc}
\toprule
\multirow{3}{*}{Model}
& \multicolumn{5}{c}{IOCD}
& \multicolumn{5}{c}{MOCV}
& \multicolumn{5}{c}{LOOO} \\
\cmidrule(lr){2-6} \cmidrule(lr){7-11} \cmidrule(lr){12-16}
& \multicolumn{2}{c}{$T_{\mathrm{fine}}$ (9)}
& \multicolumn{2}{c}{$T_{\mathrm{coarse}}$ (3)}
& \multirow{2}{*}{\shortstack{Sig.\\wins$\uparrow$}}
& \multicolumn{2}{c}{$T_{\mathrm{fine}}$ (9)}
& \multicolumn{2}{c}{$T_{\mathrm{coarse}}$ (3)}
& \multirow{2}{*}{\shortstack{Sig.\\wins$\uparrow$}}
& \multicolumn{2}{c}{$T_{\mathrm{fine}}$ (9)}
& \multicolumn{2}{c}{$T_{\mathrm{coarse}}$ (3)}
& \multirow{2}{*}{\shortstack{Sig.\\wins$\uparrow$}} \\
\cmidrule(lr){2-3} \cmidrule(lr){4-5}
\cmidrule(lr){7-8} \cmidrule(lr){9-10}
\cmidrule(lr){12-13} \cmidrule(lr){14-15}
& F1 & AUROC & F1 & AUROC &
& F1 & AUROC & F1 & AUROC &
& F1 & AUROC & F1 & AUROC & \\
\bottomrule
\end{tabular}
}
\end{table*}

\subsubsection{Within-Section Evaluation of Cell Representation Advantage}
\label{sec:within_slide_results}

Under the IS protocol, the \emph{Nuc} readout consistently outperformed \emph{Mean} across all 30 foundation models and both cell-type taxonomies (Table~\ref{tab:is_20x_cellgain}). To quantify this difference, let $F_{m,b}^{r,q,t,\pi}$ denote the Macro-F1 score (expressed in percentage points) of model $m$ on evaluation unit $b$ at magnification $r$, using readout $q$, taxonomy $t$, and protocol $\pi$. Focusing on the $20\times$ magnification under IS, we first compute the matched performance difference between the nucleus-anchored and patch-level representations, followed by a hierarchical aggregation:
\begin{equation}
\begin{aligned}
\Delta_{m,b,t}
&=
F_{m,b}^{20\times,\mathrm{Nuc},t,\mathrm{IS}}
-
F_{m,b}^{20\times,\mathrm{Mean},t,\mathrm{IS}},\\
C_{m,o}
&=
\frac{1}{\lvert\mathcal{D}_{m,o}\rvert}
\sum_{(b,t)\in\mathcal{D}_{m,o}}
\Delta_{m,b,t},\\
\mathrm{CRA}_{m}
&=
\frac{1}{\lvert\mathcal{O}_{m}\rvert}
\sum_{o\in\mathcal{O}_{m}} C_{m,o}.
\end{aligned}
\label{eq:cra}
\end{equation}

Here, $\mathcal{D}_{m,o}$ represents all valid matched evaluation-unit--taxonomy pairs from both $T_{\mathrm{fine}}$ and $T_{\mathrm{coarse}}$ for a given organ $o$, while $\mathcal{O}_{m}$ denotes the set of evaluated organs possessing at least one valid pair. By subtracting \emph{Mean} from \emph{Nuc} prior to aggregation, CRA first averages these paired differences within each organ, and then assigns equal weight to all organs. Finally, the reported standard deviation for CRA is simply the sample standard deviation computed across the intermediate organ-level averages $C_{m,o}$.

The magnitude of CRA nevertheless varied substantially across models. H-Optimus-1 achieved the highest CRA (\(10.14\pm6.05\) \%) and significantly outperformed all 29 competing models, demonstrating that CRA can further distinguish the cell-level representation capability of different encoders. Absolute \emph{Nuc} performance and CRA therefore provide complementary measurements: the former quantifies the total amount of linearly decodable cell-type information, whereas the latter quantifies the specific advantage of nucleus-anchored representation over patch-level mean pooling. 

\subsubsection{Cross-Domain Evaluation of Cell Representation Transferability}
\label{sec:cross_slide_results}

We evaluated the transferability of nucleus-anchored representations at $20\times$ under IOCD, MOCV, and LOOO. IOCD tests transfer to a held-out tissue section within a known organ, MOCV evaluates held-out sections after multi-organ training, and LOOO transfers to an unseen organ. Because these protocols differ in training composition, test domains, and aggregation units, their absolute scores are interpreted within rather than directly across protocols.
To summarize transferability across the three cross-domain protocols (IOCD, MOCV, and LOOO), we define Cell Representation Transferability (CRT). Let $W_m^{\pi}$ denote the significant-win count of model $m$ under protocol $\pi$ for a fixed reference configuration (specifically, the \emph{Nuc} readout at $20\times$ under the fine-grained taxonomy $T_{\mathrm{fine}}$). CRT is defined as the average of these win counts:
\begin{equation}
\mathrm{CRT}_{m}
=
\frac{1}{3}
\left(
W_m^{\mathrm{IOCD}}
+
W_m^{\mathrm{MOCV}}
+
W_m^{\mathrm{LOOO}}
\right).
\label{eq:crt}
\end{equation}
CRT ranges from $0$ to $29$, with higher values indicating more consistent panel-relative superiority across diverse cross-domain settings. It is important to note that CRT is not a percentage, probability, or absolute transfer-performance measure; rather, its value inherently depends on the evaluated model panel, the available matched evaluation units, and statistical power.

SEAL(UNI2), UNI2, H-Optimus-0, and H-Optimus-1 formed the leading group across protocols (Table~\ref{tab:cross}). UNI2 achieved the highest fine-grained Macro-F1 under IOCD (28.85), SEAL(UNI2) under MOCV (29.14), and H-Optimus-0 under LOOO (23.96). For the coarse-grained taxonomy, H-Optimus-0 led IOCD (49.57), whereas H-Optimus-1 led MOCV (60.68) and LOOO (54.59). Macro-AUROC showed a broadly consistent leading group.

SEAL(UNI2) and UNI2 achieved the highest CRT of 26.33, followed by H-Optimus-1 at 25.33 and H-Optimus-0 at 24.33. Within-section CRA and cross-domain CRT were not fully aligned: Virchow2 ranked second in CRA but obtained a lower CRT of 16.33, whereas UNI2 remained consistently competitive across all three transfer protocols. H-Optimus-1 performed strongly on both dimensions. CRA and CRT therefore capture complementary properties of cell-level representations: positional advantage within tissue sections and transferability across domains.

\section{Discussion and Conclusion}
\label{sec:discussion_conclusion}

CellPath-Bench provides a controlled framework for evaluating cell-type information encoded by frozen pathology foundation models. By aligning WSI feature maps with registered nuclear locations and applying standardized linear probes, it enables comparison across taxonomic granularities, spatial readouts, image resolutions, and within- and cross-domain settings.

The benchmark highlights that cell-level representation capability is multidimensional and cannot be adequately summarized by a single score. \emph{Nuc} Macro-F1 measures linearly accessible cell-type information, while CRA and CRT characterize nucleus-anchored representation advantage and cross-domain competitiveness, respectively. Together, these measures provide complementary views of local cellular information and its generalization.

CellPath-Bench remains limited to frozen linear probing and discrete cell-type classification, while CRT depends on the evaluated model panel and statistical power. 
Future extensions may incorporate continuous cell states, molecular phenotypes, and additional cell-level tasks. Overall, CellPath-Bench establishes a scalable framework for characterizing where cell-type information is encoded in pathology foundation models and how reliably it transfers across biological domains.

\bibliography{references}


\clearpage
\onecolumn

\appendix

\section{A. Evaluated Foundation Models}

\begin{table}[h]
    \centering
    \setlength{\belowcaptionskip}{8pt}
    \makeatletter
    \renewcommand{\fnum@table}{\normalsize\tablename~\thetable}
    \makeatother
    \caption{\normalsize
      The 30 foundation models evaluated in CellPath-Bench.
    }
    \label{tab:evaluated_models}
    
    \begingroup
    \scriptsize
    \setlength{\tabcolsep}{1.5pt}
    \renewcommand{\arraystretch}{1.00}
    
\begin{tabular}{
    @{}
    >{\centering\arraybackslash}p{0.035\textwidth}
    >{\raggedright\arraybackslash}p{0.210\textwidth}
    >{\raggedright\arraybackslash}p{0.155\textwidth}
    >{\centering\arraybackslash}p{0.100\textwidth}
    >{\centering\arraybackslash}p{0.075\textwidth}
    >{\raggedright\arraybackslash}p{0.140\textwidth}
    >{\centering\arraybackslash}p{0.070\textwidth}
    >{\raggedleft\arraybackslash}p{0.065\textwidth}
    >{\raggedleft\arraybackslash}p{0.075\textwidth}
    @{}
    }
    
    \toprule
    \textbf{Type}
    & \textbf{Model}
    & \mbox{\textbf{Pretraining Method}}
    & \mbox{\textbf{Pretrain Mag}}
    & \mbox{\textbf{Pretrain Res}}
    & \mbox{\textbf{Model Architecture}}
    & \mbox{\textbf{Input size}}
    & \mbox{\textbf{Params.}}
    & \mbox{\textbf{Pre. WSI}} \\
    \midrule
    
    \multirow[t]{20}{*}{PV}
    & H-Optimus-1
    & DINOv2
    & 20x
    & 224
    & ViT-G/14
    & 224
    & 1100M
    & 1000K \\
    
    & H-Optimus-0
    & DINOv2
    & 20x
    & 224
    & ViT-G/14
    & 224
    & 1100M
    & 500K \\
    
    & Prov-GigaPath
    & DINOv2
    & 20x
    & 224
    & ViT-G/14
    & 224
    & 1100M
    & 171K \\
    
    & Virchow2
    & DINOv2
    & (5,10,20,40)x
    & 224
    & ViT-H/14
    & 224
    & 632M
    & 3100K \\
    
    & Virchow
    & DINOv2
    & 20x
    & 224
    & ViT-H/14
    & 224
    & 632M
    & 1500K \\
    
    & UNI2
    & DINOv2
    & Unknown
    & 224
    & ViT-H/14
    & 224
    & 632M
    & 350K \\
    
    & GPFM
    & --
    & Unknown
    & 224
    & ViT-L/14
    & 224
    & 307M
    & 72K \\
\bottomrule
    
    \end{tabular}
    
    \endgroup
    \end{table}

\newpage
\section{B. Dataset Composition and Cell-Type Annotation}

\providecommand{\tenxdef}[2]{%
  \expandafter\gdef\csname tenxurl@#1\endcsname{#2}}
\tenxdef{1}{https://www.10xgenomics.com/products/xenium-in-situ/preview-dataset-human-breast}
\tenxdef{2}{https://www.10xgenomics.com/datasets/human-skin-preview-data-xenium-human-skin-gene-expression-panel-add-on-1-standard}
\tenxdef{3}{https://www.10xgenomics.com/datasets/ffpe-human-colorectal-cancer-data-with-human-immuno-oncology-profiling-panel-and-custom-add-on-1-standard}
\tenxdef{4}{https://www.10xgenomics.com/datasets/human-liver-data-xenium-human-multi-tissue-and-cancer-panel-1-standard}
\tenxdef{5}{https://www.10xgenomics.com/datasets/human-heart-data-xenium-human-multi-tissue-and-cancer-panel-1-standard}
\tenxdef{6}{https://www.10xgenomics.com/datasets/human-skin-preview-data-xenium-human-skin-gene-expression-panel-1-standard}
\tenxdef{7}{https://www.10xgenomics.com/datasets/human-tonsil-data-xenium-human-multi-tissue-and-cancer-panel-1-standard}
\tenxdef{8}{https://www.10xgenomics.com/datasets/human-colon-preview-data-xenium-human-colon-gene-expression-panel-1-standard}
\tenxdef{9}{https://www.10xgenomics.com/datasets/ffpe-human-lung-cancer-data-with-human-immuno-oncology-profiling-panel-and-custom-add-on-1-standard}
\tenxdef{10}{https://www.10xgenomics.com/datasets/ffpe-human-ovarian-cancer-data-with-human-immuno-oncology-profiling-panel-and-custom-add-on-1-standard}
\tenxdef{11}{https://www.10xgenomics.com/datasets/human-skin-data-xenium-human-multi-tissue-and-cancer-panel-1-standard}
\tenxdef{12}{https://www.10xgenomics.com/datasets/human-bone-and-bone-marrow-data-with-custom-add-on-panel-1-standard}
\tenxdef{13}{https://www.10xgenomics.com/datasets/ffpe-human-ductal-adenocarcinoma-data-with-human-immuno-oncology-profiling-panel-1-standard}
\tenxdef{14}{https://www.10xgenomics.com/datasets/pancreatic-cancer-with-xenium-human-multi-tissue-and-cancer-panel-1-standard}
\tenxdef{15}{https://www.10xgenomics.com/datasets/xenium-prime-ffpe-human-ovarian-cancer}
\tenxdef{16}{https://www.10xgenomics.com/datasets/xenium-ffpe-human-breast-biomarkers}
\tenxdef{17}{https://www.10xgenomics.com/datasets/human-kidney-preview-data-xenium-human-multi-tissue-and-cancer-panel-1-standard}
\tenxdef{18}{https://www.10xgenomics.com/datasets/xenium-protein-ffpe-human-renal-carcinoma}
\tenxdef{19}{https://www.10xgenomics.com/datasets/xenium-prime-ffpe-human-prostate}
\tenxdef{20}{https://www.10xgenomics.com/datasets/xenium-prime-ffpe-human-cervical-cancer}
\tenxdef{21}{https://www.10xgenomics.com/datasets/ffpe-human-pancreas-with-xenium-multimodal-cell-segmentation-1-standard}
\tenxdef{22}{https://www.10xgenomics.com/datasets/preview-data-ffpe-human-lung-cancer-with-xenium-multimodal-cell-segmentation-1-standard}
\tenxdef{23}{https://www.10xgenomics.com/datasets/xenium-prime-ffpe-human-skin}
\tenxdef{24}{https://www.10xgenomics.com/datasets/xenium-human-lung-cancer-post-xenium-technote}
\tenxdef{25}{https://www.10xgenomics.com/datasets/preview-data-xenium-prime-gene-expression}
\tenxdef{26}{https://www.10xgenomics.com/datasets/xenium-prime-ffpe-human-breast-cancer}
\tenxdef{27}{https://www.10xgenomics.com/datasets/ffpe-human-breast-with-pre-designed-panel-1-standard}
\tenxdef{28}{https://www.10xgenomics.com/datasets/ffpe-human-brain-cancer-data-with-human-immuno-oncology-profiling-panel-and-custom-add-on-1-standard}
\tenxdef{29}{https://www.10xgenomics.com/datasets/ffpe-human-breast-with-custom-add-on-panel-1-standard}
\tenxdef{30}{https://www.10xgenomics.com/datasets/ffpe-human-breast-using-the-entire-sample-area-1-standard}
\tenxdef{31}{https://www.10xgenomics.com/datasets/xenium-comparison-fresh-frozen-human-ovarian-cancer}
\tenxdef{32}{https://www.10xgenomics.com/datasets/xenium-prime-fresh-frozen-human-ovary}

\makeatletter
\providecommand{\tenxlink}[1]{\textup{(%
  \@ifundefined{href}{url~#1}%
  {\expandafter\href\expandafter{\csname tenxurl@#1\endcsname}{url~#1}})}}
\makeatother

\begingroup
\setlength{\LTleft}{0pt}
\setlength{\LTright}{0pt}
\LTcapwidth=\textwidth
\scriptsize
\setlength{\tabcolsep}{0.1pt}
\renewcommand{\arraystretch}{0.92}
\makeatletter
\renewcommand{\fnum@table}{\normalsize\tablename~\thetable}
\makeatother

\begin{longtable}{@{\extracolsep{\fill}} c l l l r c r @{}}
\caption{\normalsize
The full collection comprises 52 Xenium slides.
The 51 slides retaining at least two Coarse classes after label mapping
are ranked in descending order by \textbf{AGR-BAcc}.
Ranks are computed from the original full-precision values, whereas
AGR-BAcc is displayed to three decimal places.
The dataset \emph{Xenium V1 hBoneMarrow acute lymphoid leukemia section}
is shown last for completeness but is excluded from ranking because its
labels collapse to a single Coarse class after mapping
($3\rightarrow1\rightarrow1$), for which balanced accuracy is undefined.
\textbf{Cells} denotes the number of segmented cells in each slide.
\textbf{Cell-type mapping} reports
Cell type $\rightarrow$ Fine $\rightarrow$ Coarse,
where Cell type denotes the number of molecular cell-type labels present
in the slide from the harmonized 68-cell-type ontology, and Fine and
Coarse denote the nine-class and three-class taxonomies, respectively.
The horizontal rule after rank 25 separates the final CellPath-Bench
test panel from the remaining candidate slides.
A parenthesized marker \textup{(url~$n$)} refers to entry $n$ of the numbered
list of official 10x Genomics dataset pages that follows the table; slides
released on the same page share a number.
}
\label{tab:xenium52_agr_bacc}\\
\toprule
Rank & Dataset & Organ & Status & Cells & Cell mapping & AGR-BAcc \\
\midrule
\endfirsthead
\multicolumn{7}{l}{\tablename~\thetable\ (continued)}\\
\toprule
Rank & Dataset & Organ & Status & Cells & Cell mapping & AGR-BAcc \\
\midrule
\endhead
\midrule
\multicolumn{7}{r}{Continued on next page}\\
\endfoot
\bottomrule
\endlastfoot
1
& Xenium FFPE Human Breast Cancer Rep1 \tenxlink{1}
& Breast
& Breast cancer
& 167,780
& 16$\rightarrow$7$\rightarrow$3
& 0.704 \\

2
& Xeniumranger V1 hSkin Melanoma Add on FFPE \tenxlink{2}
& Skin
& Melanoma
& 87,499
& 7$\rightarrow$5$\rightarrow$3
& 0.669 \\

3
& Xenium V1 Human Colorectal Cancer Addon FFPE \tenxlink{3}
& Colon
& Colorectal cancer
& 388,175
& 19$\rightarrow$8$\rightarrow$3
& 0.664 \\

4
& Xenium V1 hLiver nondiseased section FFPE \tenxlink{4}
& Liver
& Normal
& 239,271
& 17$\rightarrow$7$\rightarrow$3
& 0.620 \\

5
& Xenium V1 hHeart nondiseased section FFPE \tenxlink{5}
& Heart
& Normal
& 26,366
& 14$\rightarrow$6$\rightarrow$2
& 0.599 \\

6
& Xenium V1 hSkin Melanoma Base FFPE \tenxlink{6}
& Skin
& Melanoma
& 106,980
& 7$\rightarrow$5$\rightarrow$3
& 0.595 \\

7
& Xenium V1 hTonsil reactive follicular hyperplasia section FFPE \tenxlink{7}
& Lymphoid
& Lymphoid hyperplasia
& 1,349,620
& 22$\rightarrow$8$\rightarrow$3
& 0.577 \\

8
& Xenium V1 hColon Cancer Add on FFPE \tenxlink{8}
& Colon
& Colorectal cancer
& 587,115
& 19$\rightarrow$8$\rightarrow$3
& 0.574 \\
\end{longtable}
\endgroup

\begingroup
\scriptsize\raggedright\setlength{\parindent}{0pt}
\smallskip
\noindent\textbf{Dataset source pages.} Numbers correspond to the \textup{(url~$n$)} markers in the table above; slides released on the same page share a number.\par
\smallskip
\makebox[1.6em][l]{1.}\url{https://www.10xgenomics.com/products/xenium-in-situ/preview-dataset-human-breast}\par
\makebox[1.6em][l]{2.}\url{https://www.10xgenomics.com/datasets/human-skin-preview-data-xenium-human-skin-gene-expression-panel-add-on-1-standard}\par
\makebox[1.6em][l]{3.}\url{https://www.10xgenomics.com/datasets/ffpe-human-colorectal-cancer-data-with-human-immuno-oncology-profiling-panel-and-custom-add-on-1-standard}\par
\makebox[1.6em][l]{4.}\url{https://www.10xgenomics.com/datasets/human-liver-data-xenium-human-multi-tissue-and-cancer-panel-1-standard}\par
\makebox[1.6em][l]{5.}\url{https://www.10xgenomics.com/datasets/human-heart-data-xenium-human-multi-tissue-and-cancer-panel-1-standard}\par
\makebox[1.6em][l]{6.}\url{https://www.10xgenomics.com/datasets/human-skin-preview-data-xenium-human-skin-gene-expression-panel-1-standard}\par
\makebox[1.6em][l]{7.}\url{https://www.10xgenomics.com/datasets/human-tonsil-data-xenium-human-multi-tissue-and-cancer-panel-1-standard}\par
\makebox[1.6em][l]{8.}\url{https://www.10xgenomics.com/datasets/human-colon-preview-data-xenium-human-colon-gene-expression-panel-1-standard}\par
\makebox[1.6em][l]{9.}\url{https://www.10xgenomics.com/datasets/ffpe-human-lung-cancer-data-with-human-immuno-oncology-profiling-panel-and-custom-add-on-1-standard}\par
\makebox[1.6em][l]{10.}\url{https://www.10xgenomics.com/datasets/ffpe-human-ovarian-cancer-data-with-human-immuno-oncology-profiling-panel-and-custom-add-on-1-standard}\par
\makebox[1.6em][l]{11.}\url{https://www.10xgenomics.com/datasets/human-skin-data-xenium-human-multi-tissue-and-cancer-panel-1-standard}\par
\makebox[1.6em][l]{12.}\url{https://www.10xgenomics.com/datasets/human-bone-and-bone-marrow-data-with-custom-add-on-panel-1-standard}\par
\makebox[1.6em][l]{13.}\url{https://www.10xgenomics.com/datasets/ffpe-human-ductal-adenocarcinoma-data-with-human-immuno-oncology-profiling-panel-1-standard}\par
\makebox[1.6em][l]{14.}\url{https://www.10xgenomics.com/datasets/pancreatic-cancer-with-xenium-human-multi-tissue-and-cancer-panel-1-standard}\par
\makebox[1.6em][l]{15.}\url{https://www.10xgenomics.com/datasets/xenium-prime-ffpe-human-ovarian-cancer}\par
\makebox[1.6em][l]{16.}\url{https://www.10xgenomics.com/datasets/xenium-ffpe-human-breast-biomarkers}\par
\makebox[1.6em][l]{17.}\url{https://www.10xgenomics.com/datasets/human-kidney-preview-data-xenium-human-multi-tissue-and-cancer-panel-1-standard}\par
\makebox[1.6em][l]{18.}\url{https://www.10xgenomics.com/datasets/xenium-protein-ffpe-human-renal-carcinoma}\par
\makebox[1.6em][l]{19.}\url{https://www.10xgenomics.com/datasets/xenium-prime-ffpe-human-prostate}\par
\makebox[1.6em][l]{20.}\url{https://www.10xgenomics.com/datasets/xenium-prime-ffpe-human-cervical-cancer}\par
\makebox[1.6em][l]{21.}\url{https://www.10xgenomics.com/datasets/ffpe-human-pancreas-with-xenium-multimodal-cell-segmentation-1-standard}\par
\makebox[1.6em][l]{22.}\url{https://www.10xgenomics.com/datasets/preview-data-ffpe-human-lung-cancer-with-xenium-multimodal-cell-segmentation-1-standard}\par
\makebox[1.6em][l]{23.}\url{https://www.10xgenomics.com/datasets/xenium-prime-ffpe-human-skin}\par
\makebox[1.6em][l]{24.}\url{https://www.10xgenomics.com/datasets/xenium-human-lung-cancer-post-xenium-technote}\par
\makebox[1.6em][l]{25.}\url{https://www.10xgenomics.com/datasets/preview-data-xenium-prime-gene-expression}\par
\makebox[1.6em][l]{26.}\url{https://www.10xgenomics.com/datasets/xenium-prime-ffpe-human-breast-cancer}\par
\makebox[1.6em][l]{27.}\url{https://www.10xgenomics.com/datasets/ffpe-human-breast-with-pre-designed-panel-1-standard}\par
\makebox[1.6em][l]{28.}\url{https://www.10xgenomics.com/datasets/ffpe-human-brain-cancer-data-with-human-immuno-oncology-profiling-panel-and-custom-add-on-1-standard}\par
\makebox[1.6em][l]{29.}\url{https://www.10xgenomics.com/datasets/ffpe-human-breast-with-custom-add-on-panel-1-standard}\par
\makebox[1.6em][l]{30.}\url{https://www.10xgenomics.com/datasets/ffpe-human-breast-using-the-entire-sample-area-1-standard}\par
\makebox[1.6em][l]{31.}\url{https://www.10xgenomics.com/datasets/xenium-comparison-fresh-frozen-human-ovarian-cancer}\par
\makebox[1.6em][l]{32.}\url{https://www.10xgenomics.com/datasets/xenium-prime-fresh-frozen-human-ovary}\par
\endgroup


\begin{figure}[H]
\centering
\includegraphics[width=0.84\textwidth]
{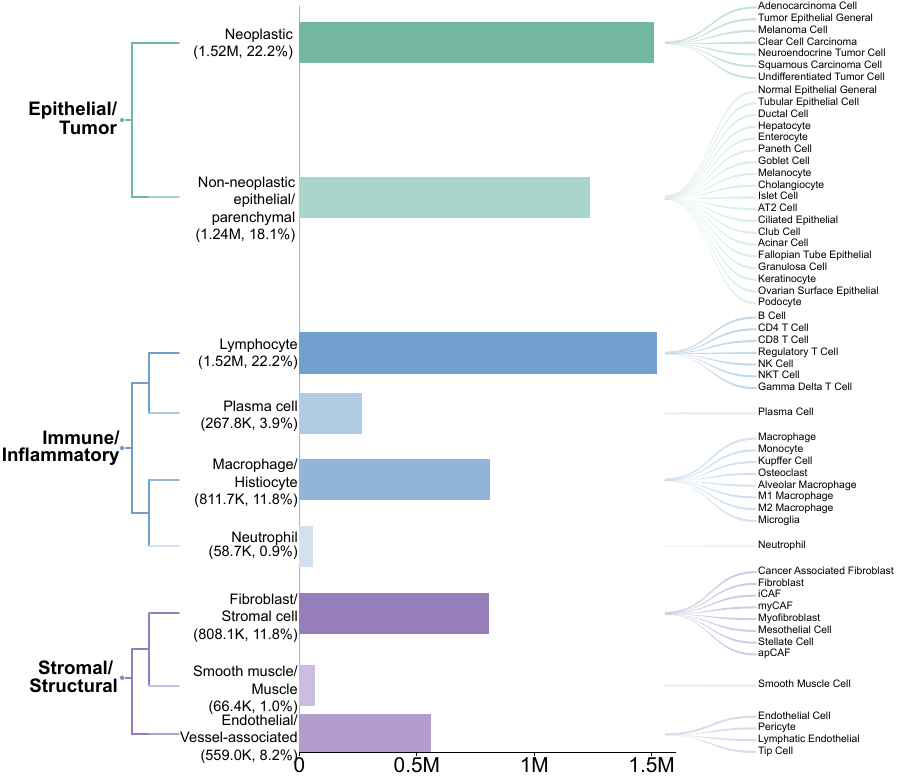}
\caption{%
Cell-type taxonomy and class abundance in the 25-slide
CellPath-Bench panel.
From left to right, the hierarchy comprises three Coarse
classes, nine Fine classes, and the corresponding mapped
cell-type labels.
Cell counts and percentages shown for each Fine class are pooled across
the full panel, and horizontal bar length represents the corresponding
number of cells.
Colors indicate the corresponding Coarse parent class.
}%
\label{fig:taxonomy_hierarchy}
\end{figure}


\begingroup

\setcitestyle{numbers,square}

\begin{bibunit}[unsrtnat]

%
%
\clearpage

\begingroup
\scriptsize
\setlength{\tabcolsep}{2.1pt}
\renewcommand{\arraystretch}{1.06}
\setlength{\LTleft}{0pt}
\setlength{\LTright}{0pt}
\LTcapwidth=\textwidth
\makeatletter
\renewcommand{\fnum@table}{\normalsize\tablename~\thetable}
\makeatother
\begin{longtable}{@{}
>{\centering\arraybackslash}p{0.025\linewidth}
>{\raggedright\arraybackslash}p{0.140\linewidth}
>{\raggedright\arraybackslash}p{0.180\linewidth}
>{\raggedright\arraybackslash}p{0.160\linewidth}
>{\raggedright\arraybackslash}p{0.150\linewidth}
>{\raggedright\arraybackslash}p{0.125\linewidth}
>{\raggedright\arraybackslash}p{0.135\linewidth}
@{}}
\caption{\normalsize
\textbf{Marker definitions for the 72 labels in the three-level annotation taxonomy.}
\textbf{Level 1}, \textbf{Level 2}, and \textbf{Cell type} correspond to the three annotation levels
defined in the finalized 72-cell-type taxonomy.
\textbf{Primary} and \textbf{secondary} markers are retained for the 68 biological cell types.
The four state or quality-control labels (necrotic, apoptotic, mitotic, and uncertain cells) have no markers or references.
}
\label{tab:celltype_markers_68}\\
\toprule
\textbf{No.} & \textbf{Level 1} & \textbf{Level 2} & \textbf{Cell type} &
\textbf{Primary markers} & \textbf{Secondary markers} & \textbf{Refs.} \\
\midrule
\endfirsthead
\multicolumn{7}{l}{\tablename~\thetable\ (continued)}\\
\toprule
\textbf{No.} & \textbf{Level 1} & \textbf{Level 2} & \textbf{Cell type} &
\textbf{Primary markers} & \textbf{Secondary markers} & \textbf{Refs.} \\
\midrule
\endhead
\midrule
\multicolumn{7}{r}{Continued on next page}\\
\endfoot
\bottomrule
\endlastfoot
1 & Epithelial & Tumor Epithelial & Tumor epithelial general & EPCAM, KRT18, KRT8 & MKI67, TOP2A & \cite{pmid42066759,pmid41895279,pmid41544627} \\
2 & Epithelial & Tumor Epithelial & Adenocarcinoma cell & CK7, CK20, MUC1, CDX2 & CEA, EMA & \cite{pmid42066759,pmid41966692,pmid40345189} \\
3 & Epithelial & Tumor Epithelial & Squamous carcinoma cell & CK5/6, p63, p40 & CK14, DSG3 & \cite{pmid41073785,pmid41529665,pmid41961948} \\
4 & Epithelial & Tumor Epithelial & Neuroendocrine tumor cell & SYN, CgA, INSM1 & -- & \cite{pmid39724105,pmid39983726,pmid41501470} \\
5 & Epithelial & Tumor Epithelial & Undifferentiated tumor cell & VIM & -- & \cite{pmid39548577} \\
6 & Epithelial & Tumor Epithelial & Clear cell carcinoma & CA9, PAX8, HIF1A & VEGFA, VHL & \cite{pmid40345189,pmid41997105} \\
7 & Epithelial & Tumor Epithelial & Melanoma cell & MLANA, SOX10, PMEL & MITF, S100B, TYR & \cite{pmid41038164,pmid41100131,pmid38759655} \\
8 & Epithelial & Normal Epithelial & Normal epithelial general & EPCAM, KRT18 & CDH1, KRT8 & \cite{pmid42185888,pmid42204317,pmid42203883} \\
\end{longtable}
\endgroup
\clearpage

\putbib[references_68_celltype_markers]

\end{bibunit}

\endgroup

\newpage
\section{C. Complete Experimental Results}

%
%

\providecommand{\ISResultsRoot}{supplementary/is_results}

%
%
%

\newcommand{\CPBISResultCaption}[5]{%
Complete IS results, \(\mathbf{S_{#1}}\) panel,
\(\mathbf{#2\times}\) magnification,
\(\mathbf{T_{\mathrm{#5}}}\) taxonomy, \textbf{#3}. See the shared legend above.%
}

%
%
\newcommand{\CPBISResultHeader}[1]{%
\toprule
Model & Nuc & Mean & CLS & N+M & N+C & N\(\Vert\)M & N\(\Vert\)C
& \(\mathrm{CRA}_{\mathrm{#1}}\) & Wins \\
\midrule
}

\begingroup
\setlength{\intextsep}{6pt}

%

\noindent
The following tables report the complete intra-section (IS) results for all 30
frozen pathology foundation models. They are organised by evaluation panel
(\(S_{25}\), \(S_{20}\), \(S_{15}\), \(S_{10}\)), then by magnification
(\(40\times\), \(20\times\), \(10\times\)), then by metric (Macro-F1,
Macro-AUROC, Macro-AUPRC), and finally by taxonomy
(\(T_{\mathrm{fine}}\), \(T_{\mathrm{coarse}}\)).

\smallskip\noindent
\textbf{Taxonomies.} \(T_{\mathrm{fine}}\) is the 9-class H\&E-aligned taxonomy and
\(T_{\mathrm{coarse}}\) is its 3-class coarsening. A class enters a given evaluation
unit only if it supplies at least 500 training and 200 test cells for
\(T_{\mathrm{fine}}\), or 100 of each for \(T_{\mathrm{coarse}}\); Macro-F1 is
averaged over the classes that survive this gate, so the number of scored classes
varies by unit.

\smallskip\noindent
\textbf{Panels.} \(S_{N}\) denotes the \(N\) highest-ranked tissue sections under the
annotation-agreement criterion. The panels are nested and cover
\(S_{25}\): 25 sections across 11 organs;
\(S_{20}\): 20 across 10;
\(S_{15}\): 15 across 9;
\(S_{10}\): 10 across 6.

\smallskip\noindent
\textbf{Readouts (column headings).} Each column is one way of turning a frozen
encoder's output into a per-cell vector.
\textbf{Nuc} samples the dense feature map at the registered nuclear centre.
\textbf{Mean} averages all spatial tokens of the patch.
\textbf{CLS} uses the architecture's native class token.
\textbf{N\(+\)M} and \textbf{N\(+\)C} are additive fusions of the nucleus readout
with the mean-pooled and the class-token readout respectively, whereas
\textbf{N\(\Vert\)M} and \textbf{N\(\Vert\)C} are the corresponding concatenative
fusions. \textbf{Wins} is the significant-wins count defined below.
A dash (--) marks a class-token-dependent readout that is structurally unavailable,
which applies to the one model in the panel without a native class token.

\smallskip\noindent
\textbf{Aggregation.} Every value is an organ-balanced mean: runs are averaged over
folds within a section, then over sections within an organ, then over organs. The
reported dispersion is the sample standard deviation across sections, so it measures
between-section variability rather than between-organ variability. All values are in
percentage points.

\smallskip\noindent
\textbf{CRA.} \(\mathrm{CRA}_{m}\) is the cellular readout advantage for metric \(m\):
the paired nucleus-minus-mean-pooling difference, computed per
model \(\times\) taxonomy \(\times\) section \(\times\) fold and pooled over both
in percentage points. It is computed from the taxonomy of the table it appears in, so
the \(T_{\mathrm{fine}}\) and \(T_{\mathrm{coarse}}\) values differ.

\smallskip\noindent
\textbf{Significant wins (Wins).} The last column counts how many of the other 29
models a given model significantly exceeds in CRA, by a two-sided Wilcoxon
signed-rank test on the paired CRA differences over their common
section \(\times\) fold units, at \(p < 0.05\) and with a
positive median difference. The maximum is therefore 29. Like CRA it uses only the
taxonomy of the table it appears in.

\smallskip\noindent
\textbf{Row order.} Models are sorted by Wins in descending order, with the unrounded
CRA as the tie-break. Every quantity in a table, including the sort key, is computed
from that table's own panel, magnification, metric and taxonomy alone, so each table
stands on its own and the two taxonomies are ordered independently.

\smallskip\noindent
\textbf{Highlighting.} Within each readout column and the CRA column the best value is
set in \textbf{bold} and the second best is \underline{underlined}; higher is better
throughout. The Wins column is not highlighted, because it is the sort key.



\begin{table}[H]
\centering
\setlength{\abovecaptionskip}{4pt}
\setlength{\belowcaptionskip}{2pt}
\scriptsize
\caption{\CPBISResultCaption{25}{40}{Macro-F1}{F1}{fine}}
\label{tab:is_s25_40x_f1_fine}

\fontsize{7}{8.6}\selectfont
\setlength{\tabcolsep}{1.5pt}
\renewcommand{\arraystretch}{1.0}

\begin{tabular*}{\textwidth}{@{\extracolsep{\fill}}lrrrrrrrrr@{}}
\CPBISResultHeader{F1}
UNI2 & \textbf{36.94{\scriptsize$\pm$12.68}} & \underline{29.29{\scriptsize$\pm$8.25}} & 15.73{\scriptsize$\pm$6.03} & \textbf{37.52{\scriptsize$\pm$12.57}} & \underline{36.94{\scriptsize$\pm$12.31}} & \textbf{37.61{\scriptsize$\pm$12.60}} & \textbf{37.44{\scriptsize$\pm$12.36}} & \textbf{7.90{\scriptsize$\pm$5.72}} & 27 \\
SEAL(UNI2) & \underline{36.93{\scriptsize$\pm$12.66}} & \textbf{29.37{\scriptsize$\pm$8.30}} & 15.83{\scriptsize$\pm$5.93} & \underline{37.51{\scriptsize$\pm$12.55}} & 36.86{\scriptsize$\pm$12.21} & \underline{37.60{\scriptsize$\pm$12.62}} & \underline{37.41{\scriptsize$\pm$12.28}} & \underline{7.87{\scriptsize$\pm$5.75}} & 27 \\
H-Optimus-1 & 36.74{\scriptsize$\pm$12.66} & 29.25{\scriptsize$\pm$8.02} & 15.93{\scriptsize$\pm$5.88} & 37.46{\scriptsize$\pm$12.48} & \textbf{37.17{\scriptsize$\pm$12.35}} & 37.38{\scriptsize$\pm$12.49} & 37.33{\scriptsize$\pm$12.22} & 7.78{\scriptsize$\pm$6.00} & 27 \\
Virchow2 & 36.27{\scriptsize$\pm$12.03} & 28.94{\scriptsize$\pm$7.91} & 16.14{\scriptsize$\pm$5.85} & 37.35{\scriptsize$\pm$11.96} & 36.74{\scriptsize$\pm$11.67} & 37.46{\scriptsize$\pm$11.95} & 37.20{\scriptsize$\pm$11.79} & 7.58{\scriptsize$\pm$5.52} & 25 \\
H0mini & 35.96{\scriptsize$\pm$12.16} & 28.89{\scriptsize$\pm$8.12} & 16.08{\scriptsize$\pm$5.94} & 36.48{\scriptsize$\pm$12.02} & 36.23{\scriptsize$\pm$11.81} & 36.64{\scriptsize$\pm$12.16} & 36.72{\scriptsize$\pm$12.04} & 7.37{\scriptsize$\pm$5.49} & 24 \\
H-Optimus-0 & 36.12{\scriptsize$\pm$12.26} & 29.17{\scriptsize$\pm$8.16} & 16.05{\scriptsize$\pm$5.76} & 36.96{\scriptsize$\pm$12.26} & 36.64{\scriptsize$\pm$12.03} & 36.81{\scriptsize$\pm$12.25} & 36.85{\scriptsize$\pm$11.99} & 7.32{\scriptsize$\pm$5.64} & 24 \\
\bottomrule
\end{tabular*}
\end{table}

\begin{table}[H]
\centering
\setlength{\abovecaptionskip}{4pt}
\setlength{\belowcaptionskip}{2pt}
\scriptsize
\caption{\CPBISResultCaption{25}{40}{Macro-F1}{F1}{coarse}}
\label{tab:is_s25_40x_f1_coarse}

\fontsize{7}{8.6}\selectfont
\setlength{\tabcolsep}{1.5pt}
\renewcommand{\arraystretch}{1.0}

\begin{tabular*}{\textwidth}{@{\extracolsep{\fill}}lrrrrrrrrr@{}}
\CPBISResultHeader{F1}
H-Optimus-1 & \textbf{56.94{\scriptsize$\pm$13.12}} & 49.99{\scriptsize$\pm$10.52} & 34.89{\scriptsize$\pm$4.80} & \textbf{57.58{\scriptsize$\pm$12.68}} & \textbf{57.69{\scriptsize$\pm$12.32}} & \textbf{57.59{\scriptsize$\pm$12.61}} & \textbf{58.05{\scriptsize$\pm$12.24}} & \textbf{7.67{\scriptsize$\pm$4.36}} & 27 \\
UNI2 & \underline{56.68{\scriptsize$\pm$13.39}} & \underline{50.12{\scriptsize$\pm$10.53}} & 34.45{\scriptsize$\pm$4.70} & 57.30{\scriptsize$\pm$12.97} & 57.01{\scriptsize$\pm$12.63} & \underline{57.48{\scriptsize$\pm$13.02}} & \underline{57.69{\scriptsize$\pm$12.59}} & 7.41{\scriptsize$\pm$4.31} & 27 \\
SEAL(UNI2) & 56.68{\scriptsize$\pm$13.38} & \textbf{50.17{\scriptsize$\pm$10.53}} & 34.68{\scriptsize$\pm$4.80} & \underline{57.35{\scriptsize$\pm$12.96}} & 56.94{\scriptsize$\pm$12.53} & 57.46{\scriptsize$\pm$12.99} & 57.69{\scriptsize$\pm$12.59} & 7.40{\scriptsize$\pm$4.31} & 27 \\
Virchow & 56.03{\scriptsize$\pm$12.78} & 49.24{\scriptsize$\pm$10.24} & \underline{35.37{\scriptsize$\pm$4.91}} & 57.02{\scriptsize$\pm$12.31} & 57.17{\scriptsize$\pm$12.00} & 57.19{\scriptsize$\pm$12.25} & 57.27{\scriptsize$\pm$11.95} & \underline{7.43{\scriptsize$\pm$4.26}} & 23 \\
Virchow2 & 56.14{\scriptsize$\pm$13.12} & 48.62{\scriptsize$\pm$10.47} & 35.03{\scriptsize$\pm$4.74} & 57.32{\scriptsize$\pm$12.40} & \underline{57.29{\scriptsize$\pm$12.02}} & 57.31{\scriptsize$\pm$12.35} & 57.56{\scriptsize$\pm$12.13} & 7.35{\scriptsize$\pm$4.36} & 23 \\
H-Optimus-0 & 56.29{\scriptsize$\pm$13.12} & 49.55{\scriptsize$\pm$10.29} & 35.00{\scriptsize$\pm$4.62} & 57.17{\scriptsize$\pm$12.38} & 57.15{\scriptsize$\pm$12.09} & 57.22{\scriptsize$\pm$12.50} & 57.56{\scriptsize$\pm$12.09} & 7.26{\scriptsize$\pm$4.22} & 23 \\
\bottomrule
\end{tabular*}
\end{table}

\begin{table}[H]
\centering
\setlength{\abovecaptionskip}{4pt}
\setlength{\belowcaptionskip}{2pt}
\scriptsize
\caption{\CPBISResultCaption{25}{40}{Macro-AUROC}{AUROC}{fine}}
\label{tab:is_s25_40x_auroc_fine}

\fontsize{7}{8.6}\selectfont
\setlength{\tabcolsep}{1.5pt}
\renewcommand{\arraystretch}{1.0}

\begin{tabular*}{\textwidth}{@{\extracolsep{\fill}}lrrrrrrrrr@{}}
\CPBISResultHeader{AUROC}
H-Optimus-1 & 75.77{\scriptsize$\pm$12.19} & 69.59{\scriptsize$\pm$9.46} & 50.00{\scriptsize$\pm$0.16} & 75.50{\scriptsize$\pm$12.33} & 75.07{\scriptsize$\pm$12.33} & 75.21{\scriptsize$\pm$12.48} & 75.22{\scriptsize$\pm$12.35} & \textbf{6.04{\scriptsize$\pm$3.79}} & 27 \\
SEAL(UNI2) & \underline{76.04{\scriptsize$\pm$12.23}} & \underline{70.04{\scriptsize$\pm$9.21}} & 50.04{\scriptsize$\pm$0.18} & \underline{76.07{\scriptsize$\pm$12.25}} & \underline{75.11{\scriptsize$\pm$12.27}} & \textbf{76.05{\scriptsize$\pm$12.35}} & \underline{75.48{\scriptsize$\pm$12.38}} & 5.79{\scriptsize$\pm$4.00} & 25 \\
UNI2 & \textbf{76.08{\scriptsize$\pm$12.19}} & \textbf{70.12{\scriptsize$\pm$9.20}} & 50.07{\scriptsize$\pm$0.20} & \textbf{76.09{\scriptsize$\pm$12.27}} & \textbf{75.32{\scriptsize$\pm$12.24}} & \underline{76.03{\scriptsize$\pm$12.37}} & \textbf{75.63{\scriptsize$\pm$12.35}} & 5.78{\scriptsize$\pm$3.95} & 25 \\
H-Optimus-0 & 75.32{\scriptsize$\pm$12.03} & 69.40{\scriptsize$\pm$9.37} & 49.98{\scriptsize$\pm$0.17} & 75.11{\scriptsize$\pm$12.26} & 74.60{\scriptsize$\pm$12.20} & 74.89{\scriptsize$\pm$12.37} & 74.72{\scriptsize$\pm$12.30} & \underline{5.88{\scriptsize$\pm$3.73}} & 24 \\
H0mini & 75.34{\scriptsize$\pm$12.13} & 69.42{\scriptsize$\pm$9.30} & 50.05{\scriptsize$\pm$0.16} & 75.06{\scriptsize$\pm$12.14} & 74.63{\scriptsize$\pm$12.09} & 74.96{\scriptsize$\pm$12.29} & 74.98{\scriptsize$\pm$12.24} & 5.81{\scriptsize$\pm$3.77} & 24 \\
Virchow2 & 75.59{\scriptsize$\pm$11.91} & 69.72{\scriptsize$\pm$9.11} & 50.03{\scriptsize$\pm$0.19} & 75.47{\scriptsize$\pm$12.12} & 74.84{\scriptsize$\pm$12.11} & 75.34{\scriptsize$\pm$12.10} & 75.01{\scriptsize$\pm$12.15} & 5.78{\scriptsize$\pm$3.84} & 23 \\
\bottomrule
\end{tabular*}
\end{table}

\begin{table}[H]
\centering
\setlength{\abovecaptionskip}{4pt}
\setlength{\belowcaptionskip}{2pt}
\scriptsize
\caption{\CPBISResultCaption{25}{40}{Macro-AUROC}{AUROC}{coarse}}
\label{tab:is_s25_40x_auroc_coarse}

\fontsize{7}{8.6}\selectfont
\setlength{\tabcolsep}{1.5pt}
\renewcommand{\arraystretch}{1.0}

\begin{tabular*}{\textwidth}{@{\extracolsep{\fill}}lrrrrrrrrr@{}}
\CPBISResultHeader{AUROC}
UNI2 & \underline{79.28{\scriptsize$\pm$11.27}} & 73.05{\scriptsize$\pm$9.60} & 50.32{\scriptsize$\pm$0.63} & \underline{79.14{\scriptsize$\pm$11.65}} & 78.59{\scriptsize$\pm$11.52} & \textbf{78.90{\scriptsize$\pm$11.98}} & \underline{78.87{\scriptsize$\pm$11.68}} & \underline{6.46{\scriptsize$\pm$3.00}} & 27 \\
SEAL(UNI2) & \textbf{79.30{\scriptsize$\pm$11.25}} & \underline{73.07{\scriptsize$\pm$9.59}} & \underline{50.44{\scriptsize$\pm$0.88}} & \textbf{79.18{\scriptsize$\pm$11.62}} & 78.16{\scriptsize$\pm$11.67} & \underline{78.89{\scriptsize$\pm$11.97}} & \textbf{78.91{\scriptsize$\pm$11.61}} & 6.45{\scriptsize$\pm$2.98} & 27 \\
H-Optimus-1 & 78.48{\scriptsize$\pm$11.97} & \textbf{73.09{\scriptsize$\pm$9.70}} & 50.05{\scriptsize$\pm$0.33} & 78.46{\scriptsize$\pm$12.01} & 78.21{\scriptsize$\pm$12.09} & 78.25{\scriptsize$\pm$12.20} & 78.06{\scriptsize$\pm$12.26} & 6.21{\scriptsize$\pm$3.20} & 27 \\
H0mini & 78.63{\scriptsize$\pm$11.49} & 72.55{\scriptsize$\pm$9.86} & 50.37{\scriptsize$\pm$0.71} & 78.34{\scriptsize$\pm$11.70} & 78.14{\scriptsize$\pm$11.45} & 78.43{\scriptsize$\pm$11.78} & 78.23{\scriptsize$\pm$11.80} & 6.18{\scriptsize$\pm$2.99} & 24 \\
H-Optimus-0 & 78.58{\scriptsize$\pm$11.55} & 71.63{\scriptsize$\pm$10.56} & 50.17{\scriptsize$\pm$0.34} & 78.60{\scriptsize$\pm$11.53} & 78.10{\scriptsize$\pm$11.64} & 78.31{\scriptsize$\pm$11.81} & 77.96{\scriptsize$\pm$11.93} & \textbf{6.60{\scriptsize$\pm$4.86}} & 23 \\
Virchow2 & 79.12{\scriptsize$\pm$11.07} & 72.75{\scriptsize$\pm$9.60} & 50.09{\scriptsize$\pm$0.37} & 78.98{\scriptsize$\pm$11.36} & \underline{78.60{\scriptsize$\pm$11.33}} & 78.43{\scriptsize$\pm$11.83} & 78.56{\scriptsize$\pm$11.53} & 6.17{\scriptsize$\pm$3.18} & 23 \\
\bottomrule
\end{tabular*}
\end{table}

\begin{table}[H]
\centering
\setlength{\abovecaptionskip}{4pt}
\setlength{\belowcaptionskip}{2pt}
\scriptsize
\caption{\CPBISResultCaption{25}{40}{Macro-AUPRC}{AUPRC}{fine}}
\label{tab:is_s25_40x_auprc_fine}

\fontsize{7}{8.6}\selectfont
\setlength{\tabcolsep}{1.5pt}
\renewcommand{\arraystretch}{1.0}

\begin{tabular*}{\textwidth}{@{\extracolsep{\fill}}lrrrrrrrrr@{}}
\CPBISResultHeader{AUPRC}
H-Optimus-1 & 39.74{\scriptsize$\pm$12.71} & 31.66{\scriptsize$\pm$8.06} & 18.91{\scriptsize$\pm$5.02} & 39.46{\scriptsize$\pm$12.63} & \underline{38.96{\scriptsize$\pm$12.51}} & 39.23{\scriptsize$\pm$12.63} & 39.09{\scriptsize$\pm$12.48} & \textbf{8.33{\scriptsize$\pm$6.21}} & 27 \\
SEAL(UNI2) & \underline{40.02{\scriptsize$\pm$12.93}} & \underline{31.90{\scriptsize$\pm$8.20}} & 18.93{\scriptsize$\pm$5.06} & \underline{40.02{\scriptsize$\pm$12.75}} & 38.77{\scriptsize$\pm$12.30} & \underline{40.00{\scriptsize$\pm$12.83}} & \underline{39.28{\scriptsize$\pm$12.46}} & \underline{8.29{\scriptsize$\pm$6.16}} & 27 \\
UNI2 & \textbf{40.04{\scriptsize$\pm$12.93}} & \textbf{31.97{\scriptsize$\pm$8.31}} & 18.94{\scriptsize$\pm$5.07} & \textbf{40.05{\scriptsize$\pm$12.77}} & \textbf{39.04{\scriptsize$\pm$12.40}} & \textbf{40.01{\scriptsize$\pm$12.85}} & \textbf{39.42{\scriptsize$\pm$12.60}} & 8.28{\scriptsize$\pm$6.13} & 27 \\
H-Optimus-0 & 39.02{\scriptsize$\pm$12.42} & 31.45{\scriptsize$\pm$8.08} & 18.90{\scriptsize$\pm$5.00} & 38.84{\scriptsize$\pm$12.39} & 38.14{\scriptsize$\pm$12.16} & 38.64{\scriptsize$\pm$12.37} & 38.36{\scriptsize$\pm$12.22} & 7.93{\scriptsize$\pm$6.05} & 24 \\
H0mini & 38.96{\scriptsize$\pm$12.35} & 31.34{\scriptsize$\pm$7.97} & 18.94{\scriptsize$\pm$5.07} & 38.61{\scriptsize$\pm$12.05} & 38.03{\scriptsize$\pm$11.93} & 38.55{\scriptsize$\pm$12.23} & 38.58{\scriptsize$\pm$12.20} & 7.86{\scriptsize$\pm$5.88} & 24 \\
Virchow2 & 39.19{\scriptsize$\pm$12.19} & 31.66{\scriptsize$\pm$8.05} & 18.93{\scriptsize$\pm$5.07} & 39.11{\scriptsize$\pm$12.18} & 38.26{\scriptsize$\pm$11.84} & 39.05{\scriptsize$\pm$12.16} & 38.56{\scriptsize$\pm$11.98} & 7.78{\scriptsize$\pm$5.73} & 24 \\
\bottomrule
\end{tabular*}
\end{table}

\begin{table}[H]
\centering
\setlength{\abovecaptionskip}{4pt}
\setlength{\belowcaptionskip}{2pt}
\scriptsize
\caption{\CPBISResultCaption{25}{40}{Macro-AUPRC}{AUPRC}{coarse}}
\label{tab:is_s25_40x_auprc_coarse}

\fontsize{7}{8.6}\selectfont
\setlength{\tabcolsep}{1.5pt}
\renewcommand{\arraystretch}{1.0}

\begin{tabular*}{\textwidth}{@{\extracolsep{\fill}}lrrrrrrrrr@{}}
\CPBISResultHeader{AUPRC}
H-Optimus-1 & 61.11{\scriptsize$\pm$19.87} & 53.64{\scriptsize$\pm$17.55} & 35.96{\scriptsize$\pm$14.57} & 60.91{\scriptsize$\pm$19.96} & \underline{60.62{\scriptsize$\pm$19.87}} & 60.67{\scriptsize$\pm$20.03} & 60.65{\scriptsize$\pm$19.93} & \textbf{8.26{\scriptsize$\pm$4.76}} & 27 \\
UNI2 & \textbf{61.32{\scriptsize$\pm$19.85}} & \underline{53.90{\scriptsize$\pm$17.53}} & 35.97{\scriptsize$\pm$14.55} & \underline{61.40{\scriptsize$\pm$19.90}} & \textbf{60.70{\scriptsize$\pm$19.72}} & \textbf{61.29{\scriptsize$\pm$20.06}} & \textbf{60.98{\scriptsize$\pm$19.91}} & \underline{8.21{\scriptsize$\pm$4.52}} & 27 \\
SEAL(UNI2) & \underline{61.32{\scriptsize$\pm$19.85}} & \textbf{53.91{\scriptsize$\pm$17.56}} & 35.97{\scriptsize$\pm$14.55} & \textbf{61.42{\scriptsize$\pm$19.89}} & 60.40{\scriptsize$\pm$19.70} & \underline{61.28{\scriptsize$\pm$20.05}} & \underline{60.88{\scriptsize$\pm$19.91}} & 8.19{\scriptsize$\pm$4.49} & 27 \\
H0mini & 60.61{\scriptsize$\pm$19.74} & 53.42{\scriptsize$\pm$17.59} & 35.97{\scriptsize$\pm$14.53} & 60.43{\scriptsize$\pm$19.72} & 59.94{\scriptsize$\pm$19.58} & 60.40{\scriptsize$\pm$19.83} & 60.35{\scriptsize$\pm$19.78} & 7.94{\scriptsize$\pm$4.57} & 25 \\
H-Optimus-0 & 60.53{\scriptsize$\pm$19.78} & 53.30{\scriptsize$\pm$17.34} & 35.99{\scriptsize$\pm$14.51} & 60.45{\scriptsize$\pm$19.77} & 60.08{\scriptsize$\pm$19.68} & 60.28{\scriptsize$\pm$19.86} & 60.13{\scriptsize$\pm$19.80} & 7.92{\scriptsize$\pm$4.58} & 25 \\
Virchow2 & 60.65{\scriptsize$\pm$19.61} & 53.72{\scriptsize$\pm$17.52} & 35.98{\scriptsize$\pm$14.59} & 60.65{\scriptsize$\pm$19.68} & 60.12{\scriptsize$\pm$19.60} & 60.55{\scriptsize$\pm$19.74} & 60.28{\scriptsize$\pm$19.69} & 7.67{\scriptsize$\pm$4.47} & 22 \\
\bottomrule
\end{tabular*}
\end{table}


\begin{table}[H]
\centering
\setlength{\abovecaptionskip}{4pt}
\setlength{\belowcaptionskip}{2pt}
\scriptsize
\caption{\CPBISResultCaption{25}{20}{Macro-F1}{F1}{fine}}
\label{tab:is_s25_20x_f1_fine}

\fontsize{7}{8.6}\selectfont
\setlength{\tabcolsep}{1.5pt}
\renewcommand{\arraystretch}{1.0}

\begin{tabular*}{\textwidth}{@{\extracolsep{\fill}}lrrrrrrrrr@{}}
\CPBISResultHeader{F1}
H-Optimus-1 & \textbf{36.73{\scriptsize$\pm$12.19}} & \textbf{26.95{\scriptsize$\pm$7.63}} & 15.86{\scriptsize$\pm$5.77} & \textbf{37.29{\scriptsize$\pm$12.10}} & \textbf{37.22{\scriptsize$\pm$12.06}} & \textbf{37.22{\scriptsize$\pm$12.17}} & \textbf{37.38{\scriptsize$\pm$11.89}} & \textbf{10.10{\scriptsize$\pm$6.43}} & 28 \\
UNI2 & \underline{36.41{\scriptsize$\pm$12.08}} & 26.60{\scriptsize$\pm$7.32} & 15.58{\scriptsize$\pm$5.85} & 36.77{\scriptsize$\pm$11.90} & 36.71{\scriptsize$\pm$11.84} & 36.87{\scriptsize$\pm$11.99} & 36.98{\scriptsize$\pm$11.74} & 9.95{\scriptsize$\pm$6.32} & 26 \\
Virchow2 & 36.21{\scriptsize$\pm$11.62} & 26.45{\scriptsize$\pm$7.26} & 16.11{\scriptsize$\pm$5.88} & 36.66{\scriptsize$\pm$11.56} & 36.54{\scriptsize$\pm$11.39} & 36.86{\scriptsize$\pm$11.66} & 36.91{\scriptsize$\pm$11.56} & \underline{9.99{\scriptsize$\pm$6.15}} & 25 \\
H-Optimus-0 & 36.38{\scriptsize$\pm$11.94} & \underline{26.80{\scriptsize$\pm$7.57}} & 16.09{\scriptsize$\pm$5.84} & \underline{36.99{\scriptsize$\pm$11.82}} & \underline{36.93{\scriptsize$\pm$11.71}} & \underline{36.97{\scriptsize$\pm$11.95}} & \underline{37.12{\scriptsize$\pm$11.69}} & 9.91{\scriptsize$\pm$6.16} & 25 \\
SEAL(UNI2) & 36.41{\scriptsize$\pm$12.09} & 26.72{\scriptsize$\pm$7.38} & 15.60{\scriptsize$\pm$5.86} & 36.82{\scriptsize$\pm$11.93} & 36.66{\scriptsize$\pm$11.77} & 36.91{\scriptsize$\pm$12.02} & 36.94{\scriptsize$\pm$11.79} & 9.85{\scriptsize$\pm$6.35} & 25 \\
Virchow & 35.91{\scriptsize$\pm$11.82} & 26.64{\scriptsize$\pm$7.56} & 16.04{\scriptsize$\pm$5.85} & 36.70{\scriptsize$\pm$11.57} & 36.76{\scriptsize$\pm$11.54} & 36.78{\scriptsize$\pm$11.77} & 36.81{\scriptsize$\pm$11.47} & 9.60{\scriptsize$\pm$6.09} & 24 \\
\bottomrule
\end{tabular*}
\end{table}

\begin{table}[H]
\centering
\setlength{\abovecaptionskip}{4pt}
\setlength{\belowcaptionskip}{2pt}
\scriptsize
\caption{\CPBISResultCaption{25}{20}{Macro-F1}{F1}{coarse}}
\label{tab:is_s25_20x_f1_coarse}

\fontsize{7}{8.6}\selectfont
\setlength{\tabcolsep}{1.5pt}
\renewcommand{\arraystretch}{1.0}

\begin{tabular*}{\textwidth}{@{\extracolsep{\fill}}lrrrrrrrrr@{}}
\CPBISResultHeader{F1}
H-Optimus-1 & \textbf{56.88{\scriptsize$\pm$13.19}} & \textbf{47.81{\scriptsize$\pm$10.11}} & 34.92{\scriptsize$\pm$4.65} & \textbf{57.53{\scriptsize$\pm$12.72}} & \textbf{57.81{\scriptsize$\pm$12.44}} & \textbf{57.25{\scriptsize$\pm$12.97}} & \textbf{57.76{\scriptsize$\pm$12.41}} & \underline{10.18{\scriptsize$\pm$5.63}} & 29 \\
Virchow2 & 55.90{\scriptsize$\pm$13.14} & 45.97{\scriptsize$\pm$9.86} & 35.06{\scriptsize$\pm$4.72} & 56.98{\scriptsize$\pm$12.33} & 57.03{\scriptsize$\pm$12.19} & 56.86{\scriptsize$\pm$12.35} & 57.16{\scriptsize$\pm$12.07} & \textbf{10.27{\scriptsize$\pm$5.47}} & 24 \\
Virchow & 56.07{\scriptsize$\pm$13.15} & 46.96{\scriptsize$\pm$9.83} & 35.26{\scriptsize$\pm$4.85} & 57.06{\scriptsize$\pm$12.30} & 57.11{\scriptsize$\pm$12.25} & 57.04{\scriptsize$\pm$12.38} & 57.39{\scriptsize$\pm$12.12} & 10.06{\scriptsize$\pm$5.36} & 24 \\
H-Optimus-0 & \underline{56.66{\scriptsize$\pm$13.17}} & \underline{47.68{\scriptsize$\pm$9.93}} & 34.79{\scriptsize$\pm$4.38} & \underline{57.31{\scriptsize$\pm$12.53}} & \underline{57.54{\scriptsize$\pm$12.24}} & \underline{57.14{\scriptsize$\pm$12.83}} & \underline{57.64{\scriptsize$\pm$12.32}} & 10.01{\scriptsize$\pm$5.60} & 24 \\
SEAL(UNI2) & 56.40{\scriptsize$\pm$13.33} & 47.44{\scriptsize$\pm$10.16} & 34.64{\scriptsize$\pm$4.76} & 56.70{\scriptsize$\pm$12.95} & 56.90{\scriptsize$\pm$12.46} & 56.87{\scriptsize$\pm$12.94} & 57.17{\scriptsize$\pm$12.44} & 9.94{\scriptsize$\pm$5.71} & 24 \\
UNI2 & 56.40{\scriptsize$\pm$13.32} & 47.42{\scriptsize$\pm$10.16} & 34.43{\scriptsize$\pm$4.76} & 56.75{\scriptsize$\pm$12.94} & 56.95{\scriptsize$\pm$12.49} & 56.77{\scriptsize$\pm$12.95} & 57.29{\scriptsize$\pm$12.36} & 9.93{\scriptsize$\pm$5.72} & 24 \\
\bottomrule
\end{tabular*}
\end{table}

\begin{table}[H]
\centering
\setlength{\abovecaptionskip}{4pt}
\setlength{\belowcaptionskip}{2pt}
\scriptsize
\caption{\CPBISResultCaption{25}{20}{Macro-AUROC}{AUROC}{fine}}
\label{tab:is_s25_20x_auroc_fine}

\fontsize{7}{8.6}\selectfont
\setlength{\tabcolsep}{1.5pt}
\renewcommand{\arraystretch}{1.0}

\begin{tabular*}{\textwidth}{@{\extracolsep{\fill}}lrrrrrrrrr@{}}
\CPBISResultHeader{AUROC}
H-Optimus-1 & \textbf{75.77{\scriptsize$\pm$12.14}} & 67.58{\scriptsize$\pm$8.82} & 49.98{\scriptsize$\pm$0.20} & 75.39{\scriptsize$\pm$12.33} & \textbf{75.08{\scriptsize$\pm$12.43}} & 75.12{\scriptsize$\pm$12.58} & \underline{75.15{\scriptsize$\pm$12.36}} & \textbf{8.16{\scriptsize$\pm$4.63}} & 29 \\
H-Optimus-0 & 75.52{\scriptsize$\pm$12.08} & 67.66{\scriptsize$\pm$8.61} & 49.97{\scriptsize$\pm$0.22} & 75.19{\scriptsize$\pm$12.24} & 74.98{\scriptsize$\pm$12.25} & 74.94{\scriptsize$\pm$12.54} & 75.08{\scriptsize$\pm$12.32} & \underline{7.92{\scriptsize$\pm$4.65}} & 25 \\
Virchow2 & 75.28{\scriptsize$\pm$11.77} & 67.44{\scriptsize$\pm$8.17} & 49.99{\scriptsize$\pm$0.17} & 74.71{\scriptsize$\pm$12.14} & 74.36{\scriptsize$\pm$12.10} & 74.64{\scriptsize$\pm$12.30} & 74.60{\scriptsize$\pm$12.15} & 7.90{\scriptsize$\pm$4.83} & 25 \\
UNI2 & \underline{75.57{\scriptsize$\pm$12.14}} & \underline{67.86{\scriptsize$\pm$8.38}} & 50.02{\scriptsize$\pm$0.18} & \textbf{75.49{\scriptsize$\pm$12.27}} & \underline{75.00{\scriptsize$\pm$12.27}} & \underline{75.31{\scriptsize$\pm$12.57}} & \textbf{75.20{\scriptsize$\pm$12.29}} & 7.71{\scriptsize$\pm$4.92} & 25 \\
SEAL(UNI2) & 75.56{\scriptsize$\pm$12.15} & \textbf{67.90{\scriptsize$\pm$8.32}} & 49.99{\scriptsize$\pm$0.18} & \underline{75.47{\scriptsize$\pm$12.30}} & 74.88{\scriptsize$\pm$12.19} & \textbf{75.36{\scriptsize$\pm$12.56}} & 75.13{\scriptsize$\pm$12.39} & 7.66{\scriptsize$\pm$4.93} & 24 \\
H0mini & 75.01{\scriptsize$\pm$12.07} & 67.45{\scriptsize$\pm$8.35} & \underline{50.27{\scriptsize$\pm$2.30}} & 74.59{\scriptsize$\pm$12.08} & 74.40{\scriptsize$\pm$12.00} & 74.38{\scriptsize$\pm$12.32} & 74.55{\scriptsize$\pm$12.25} & 7.64{\scriptsize$\pm$4.81} & 23 \\
\bottomrule
\end{tabular*}
\end{table}

\begin{table}[H]
\centering
\setlength{\abovecaptionskip}{4pt}
\setlength{\belowcaptionskip}{2pt}
\scriptsize
\caption{\CPBISResultCaption{25}{20}{Macro-AUROC}{AUROC}{coarse}}
\label{tab:is_s25_20x_auroc_coarse}

\fontsize{7}{8.6}\selectfont
\setlength{\tabcolsep}{1.5pt}
\renewcommand{\arraystretch}{1.0}

\begin{tabular*}{\textwidth}{@{\extracolsep{\fill}}lrrrrrrrrr@{}}
\CPBISResultHeader{AUROC}
H-Optimus-1 & 78.47{\scriptsize$\pm$12.17} & 69.90{\scriptsize$\pm$9.59} & 50.23{\scriptsize$\pm$0.59} & 78.32{\scriptsize$\pm$12.35} & 78.27{\scriptsize$\pm$12.27} & 77.95{\scriptsize$\pm$12.64} & 78.03{\scriptsize$\pm$12.29} & \textbf{9.37{\scriptsize$\pm$5.07}} & 28 \\
H-Optimus-0 & 78.35{\scriptsize$\pm$12.18} & 69.35{\scriptsize$\pm$10.16} & 50.17{\scriptsize$\pm$0.60} & 78.03{\scriptsize$\pm$12.52} & 78.34{\scriptsize$\pm$11.92} & 77.68{\scriptsize$\pm$12.80} & 77.93{\scriptsize$\pm$12.36} & \underline{9.36{\scriptsize$\pm$4.88}} & 28 \\
SEAL(UNI2) & 78.89{\scriptsize$\pm$11.36} & \underline{70.68{\scriptsize$\pm$9.03}} & 50.25{\scriptsize$\pm$0.70} & \textbf{78.83{\scriptsize$\pm$11.82}} & 78.33{\scriptsize$\pm$11.65} & \underline{78.55{\scriptsize$\pm$12.08}} & \textbf{78.87{\scriptsize$\pm$11.41}} & 8.82{\scriptsize$\pm$4.04} & 24 \\
Virchow2 & \textbf{78.99{\scriptsize$\pm$11.19}} & 70.51{\scriptsize$\pm$8.76} & 50.35{\scriptsize$\pm$0.90} & 78.48{\scriptsize$\pm$11.55} & \textbf{78.90{\scriptsize$\pm$11.12}} & 77.81{\scriptsize$\pm$12.03} & 78.19{\scriptsize$\pm$11.73} & 8.86{\scriptsize$\pm$4.21} & 23 \\
UNI2 & \underline{78.92{\scriptsize$\pm$11.37}} & \textbf{70.70{\scriptsize$\pm$9.00}} & 50.24{\scriptsize$\pm$0.77} & \underline{78.80{\scriptsize$\pm$11.82}} & \underline{78.49{\scriptsize$\pm$11.58}} & \textbf{78.58{\scriptsize$\pm$12.07}} & \underline{78.73{\scriptsize$\pm$11.53}} & 8.79{\scriptsize$\pm$4.10} & 23 \\
H0mini & 78.18{\scriptsize$\pm$11.85} & 70.38{\scriptsize$\pm$9.09} & \underline{50.70{\scriptsize$\pm$2.63}} & 77.78{\scriptsize$\pm$12.08} & 78.19{\scriptsize$\pm$11.36} & 77.56{\scriptsize$\pm$12.24} & 77.86{\scriptsize$\pm$12.02} & 8.73{\scriptsize$\pm$4.26} & 23 \\
\bottomrule
\end{tabular*}
\end{table}

\begin{table}[H]
\centering
\setlength{\abovecaptionskip}{4pt}
\setlength{\belowcaptionskip}{2pt}
\scriptsize
\caption{\CPBISResultCaption{25}{20}{Macro-AUPRC}{AUPRC}{fine}}
\label{tab:is_s25_20x_auprc_fine}

\fontsize{7}{8.6}\selectfont
\setlength{\tabcolsep}{1.5pt}
\renewcommand{\arraystretch}{1.0}

\begin{tabular*}{\textwidth}{@{\extracolsep{\fill}}lrrrrrrrrr@{}}
\CPBISResultHeader{AUPRC}
H-Optimus-1 & \textbf{39.55{\scriptsize$\pm$12.14}} & 29.66{\scriptsize$\pm$7.65} & 18.92{\scriptsize$\pm$5.04} & \textbf{39.22{\scriptsize$\pm$12.17}} & \textbf{38.89{\scriptsize$\pm$12.05}} & 39.00{\scriptsize$\pm$12.26} & \textbf{38.91{\scriptsize$\pm$11.98}} & \textbf{10.03{\scriptsize$\pm$6.46}} & 29 \\
UNI2 & \underline{39.28{\scriptsize$\pm$12.14}} & \underline{29.71{\scriptsize$\pm$7.67}} & 18.90{\scriptsize$\pm$5.00} & \underline{39.20{\scriptsize$\pm$12.01}} & \underline{38.59{\scriptsize$\pm$11.77}} & \underline{39.10{\scriptsize$\pm$12.17}} & \underline{38.79{\scriptsize$\pm$11.79}} & \underline{9.72{\scriptsize$\pm$6.39}} & 26 \\
SEAL(UNI2) & 39.27{\scriptsize$\pm$12.14} & \textbf{29.73{\scriptsize$\pm$7.67}} & 18.90{\scriptsize$\pm$5.01} & 39.20{\scriptsize$\pm$12.04} & 38.39{\scriptsize$\pm$11.67} & \textbf{39.14{\scriptsize$\pm$12.18}} & 38.74{\scriptsize$\pm$11.83} & 9.69{\scriptsize$\pm$6.42} & 26 \\
H-Optimus-0 & 39.11{\scriptsize$\pm$11.87} & 29.62{\scriptsize$\pm$7.66} & 18.89{\scriptsize$\pm$5.01} & 38.76{\scriptsize$\pm$11.81} & 38.48{\scriptsize$\pm$11.69} & 38.65{\scriptsize$\pm$12.00} & 38.61{\scriptsize$\pm$11.78} & 9.69{\scriptsize$\pm$6.28} & 26 \\
Virchow2 & 38.69{\scriptsize$\pm$11.53} & 29.41{\scriptsize$\pm$7.39} & 18.90{\scriptsize$\pm$5.02} & 38.16{\scriptsize$\pm$11.53} & 37.72{\scriptsize$\pm$11.32} & 38.13{\scriptsize$\pm$11.62} & 37.95{\scriptsize$\pm$11.43} & 9.41{\scriptsize$\pm$6.07} & 25 \\
Virchow & 38.60{\scriptsize$\pm$11.77} & 29.56{\scriptsize$\pm$7.52} & 18.92{\scriptsize$\pm$5.03} & 38.31{\scriptsize$\pm$11.57} & 38.26{\scriptsize$\pm$11.52} & 38.35{\scriptsize$\pm$11.82} & 38.23{\scriptsize$\pm$11.64} & 9.18{\scriptsize$\pm$6.31} & 23 \\
\bottomrule
\end{tabular*}
\end{table}

\begin{table}[H]
\centering
\setlength{\abovecaptionskip}{4pt}
\setlength{\belowcaptionskip}{2pt}
\scriptsize
\caption{\CPBISResultCaption{25}{20}{Macro-AUPRC}{AUPRC}{coarse}}
\label{tab:is_s25_20x_auprc_coarse}

\fontsize{7}{8.6}\selectfont
\setlength{\tabcolsep}{1.5pt}
\renewcommand{\arraystretch}{1.0}

\begin{tabular*}{\textwidth}{@{\extracolsep{\fill}}lrrrrrrrrr@{}}
\CPBISResultHeader{AUPRC}
H-Optimus-1 & \textbf{61.16{\scriptsize$\pm$19.82}} & 50.98{\scriptsize$\pm$17.31} & 35.95{\scriptsize$\pm$14.52} & \textbf{60.90{\scriptsize$\pm$19.90}} & \textbf{60.93{\scriptsize$\pm$19.75}} & 60.48{\scriptsize$\pm$20.16} & \textbf{60.66{\scriptsize$\pm$19.83}} & \textbf{11.22{\scriptsize$\pm$6.09}} & 29 \\
H-Optimus-0 & \underline{60.98{\scriptsize$\pm$19.75}} & 51.04{\scriptsize$\pm$16.96} & 35.95{\scriptsize$\pm$14.53} & 60.69{\scriptsize$\pm$19.77} & \underline{60.62{\scriptsize$\pm$19.71}} & 60.38{\scriptsize$\pm$19.96} & 60.50{\scriptsize$\pm$19.76} & \underline{10.94{\scriptsize$\pm$5.76}} & 27 \\
SEAL(UNI2) & 60.89{\scriptsize$\pm$19.72} & \underline{51.29{\scriptsize$\pm$17.08}} & 35.95{\scriptsize$\pm$14.50} & \underline{60.86{\scriptsize$\pm$19.79}} & 60.44{\scriptsize$\pm$19.60} & \textbf{60.71{\scriptsize$\pm$19.95}} & 60.51{\scriptsize$\pm$19.68} & 10.67{\scriptsize$\pm$5.76} & 25 \\
UNI2 & 60.89{\scriptsize$\pm$19.73} & \textbf{51.31{\scriptsize$\pm$17.07}} & 35.95{\scriptsize$\pm$14.50} & 60.85{\scriptsize$\pm$19.78} & 60.58{\scriptsize$\pm$19.59} & \underline{60.70{\scriptsize$\pm$19.96}} & \underline{60.62{\scriptsize$\pm$19.68}} & 10.64{\scriptsize$\pm$5.81} & 25 \\
H0mini & 60.45{\scriptsize$\pm$19.59} & 50.89{\scriptsize$\pm$17.08} & 36.15{\scriptsize$\pm$14.61} & 60.10{\scriptsize$\pm$19.54} & 59.93{\scriptsize$\pm$19.35} & 59.91{\scriptsize$\pm$19.66} & 60.17{\scriptsize$\pm$19.59} & 10.62{\scriptsize$\pm$5.73} & 24 \\
Virchow & 60.54{\scriptsize$\pm$19.62} & 51.03{\scriptsize$\pm$17.13} & 35.96{\scriptsize$\pm$14.50} & 60.29{\scriptsize$\pm$19.60} & 60.31{\scriptsize$\pm$19.50} & 60.00{\scriptsize$\pm$19.73} & 60.20{\scriptsize$\pm$19.61} & 10.53{\scriptsize$\pm$5.96} & 23 \\
\bottomrule
\end{tabular*}
\end{table}


\begin{table}[H]
\centering
\setlength{\abovecaptionskip}{4pt}
\setlength{\belowcaptionskip}{2pt}
\scriptsize
\caption{\CPBISResultCaption{25}{10}{Macro-F1}{F1}{fine}}
\label{tab:is_s25_10x_f1_fine}

\fontsize{7}{8.6}\selectfont
\setlength{\tabcolsep}{1.5pt}
\renewcommand{\arraystretch}{1.0}

\begin{tabular*}{\textwidth}{@{\extracolsep{\fill}}lrrrrrrrrr@{}}
\CPBISResultHeader{F1}
CONCH v1.5 & \underline{33.63{\scriptsize$\pm$10.87}} & 23.42{\scriptsize$\pm$6.78} & 16.04{\scriptsize$\pm$5.85} & \underline{33.90{\scriptsize$\pm$10.31}} & \textbf{34.53{\scriptsize$\pm$10.57}} & \textbf{34.35{\scriptsize$\pm$10.75}} & \textbf{34.54{\scriptsize$\pm$10.52}} & \textbf{10.41{\scriptsize$\pm$6.48}} & 28 \\
Virchow2 & \textbf{33.71{\scriptsize$\pm$10.06}} & \textbf{24.04{\scriptsize$\pm$6.72}} & 15.94{\scriptsize$\pm$5.84} & \textbf{34.00{\scriptsize$\pm$10.00}} & \underline{34.17{\scriptsize$\pm$9.87}} & \underline{34.14{\scriptsize$\pm$9.98}} & \underline{34.31{\scriptsize$\pm$9.81}} & \underline{10.01{\scriptsize$\pm$6.09}} & 27 \\
SEAL(UNI2) & 33.47{\scriptsize$\pm$10.32} & 23.79{\scriptsize$\pm$6.76} & 15.56{\scriptsize$\pm$6.07} & 33.74{\scriptsize$\pm$10.07} & 34.09{\scriptsize$\pm$10.10} & 33.76{\scriptsize$\pm$10.33} & 34.13{\scriptsize$\pm$10.10} & 9.84{\scriptsize$\pm$5.95} & 24 \\
UNI2 & 33.46{\scriptsize$\pm$10.30} & 23.82{\scriptsize$\pm$6.63} & 15.43{\scriptsize$\pm$6.01} & 33.79{\scriptsize$\pm$10.10} & 34.04{\scriptsize$\pm$10.18} & 33.78{\scriptsize$\pm$10.33} & 34.11{\scriptsize$\pm$10.10} & 9.83{\scriptsize$\pm$6.01} & 23 \\
H-Optimus-1 & 33.31{\scriptsize$\pm$10.35} & 23.65{\scriptsize$\pm$6.59} & 15.70{\scriptsize$\pm$5.91} & 33.61{\scriptsize$\pm$10.04} & 33.94{\scriptsize$\pm$10.01} & 33.53{\scriptsize$\pm$10.24} & 33.85{\scriptsize$\pm$9.99} & 9.77{\scriptsize$\pm$5.75} & 22 \\
CONCH & 33.27{\scriptsize$\pm$10.38} & 23.67{\scriptsize$\pm$6.60} & 15.85{\scriptsize$\pm$6.01} & 33.73{\scriptsize$\pm$10.06} & 34.13{\scriptsize$\pm$10.18} & 33.80{\scriptsize$\pm$10.36} & 34.10{\scriptsize$\pm$10.20} & 9.78{\scriptsize$\pm$6.19} & 21 \\
\bottomrule
\end{tabular*}
\end{table}

\begin{table}[H]
\centering
\setlength{\abovecaptionskip}{4pt}
\setlength{\belowcaptionskip}{2pt}
\scriptsize
\caption{\CPBISResultCaption{25}{10}{Macro-F1}{F1}{coarse}}
\label{tab:is_s25_10x_f1_coarse}

\fontsize{7}{8.6}\selectfont
\setlength{\tabcolsep}{1.5pt}
\renewcommand{\arraystretch}{1.0}

\begin{tabular*}{\textwidth}{@{\extracolsep{\fill}}lrrrrrrrrr@{}}
\CPBISResultHeader{F1}
CONCH v1.5 & \underline{54.16{\scriptsize$\pm$12.90}} & 42.97{\scriptsize$\pm$9.25} & 35.01{\scriptsize$\pm$4.44} & \textbf{54.94{\scriptsize$\pm$11.61}} & \textbf{55.57{\scriptsize$\pm$11.79}} & \textbf{54.75{\scriptsize$\pm$12.09}} & \textbf{55.60{\scriptsize$\pm$11.67}} & \textbf{11.67{\scriptsize$\pm$7.23}} & 27 \\
CONCH & 53.72{\scriptsize$\pm$12.60} & 42.95{\scriptsize$\pm$8.85} & 34.90{\scriptsize$\pm$4.64} & 54.56{\scriptsize$\pm$11.66} & 55.09{\scriptsize$\pm$11.66} & 54.25{\scriptsize$\pm$12.00} & 55.11{\scriptsize$\pm$11.39} & 11.37{\scriptsize$\pm$6.71} & 26 \\
H-Optimus-1 & \textbf{54.27{\scriptsize$\pm$12.19}} & 44.05{\scriptsize$\pm$9.23} & 34.63{\scriptsize$\pm$4.34} & 54.53{\scriptsize$\pm$11.40} & \underline{55.19{\scriptsize$\pm$11.39}} & 54.38{\scriptsize$\pm$11.87} & \underline{55.22{\scriptsize$\pm$11.30}} & 11.23{\scriptsize$\pm$6.59} & 25 \\
Virchow2 & 53.98{\scriptsize$\pm$12.18} & 42.59{\scriptsize$\pm$9.06} & 35.20{\scriptsize$\pm$4.78} & 54.22{\scriptsize$\pm$11.51} & 55.01{\scriptsize$\pm$11.42} & 54.32{\scriptsize$\pm$11.78} & 54.95{\scriptsize$\pm$11.31} & \underline{11.58{\scriptsize$\pm$6.05}} & 22 \\
Kaiko ViT-B/8 & 53.45{\scriptsize$\pm$12.65} & 43.24{\scriptsize$\pm$8.95} & 35.22{\scriptsize$\pm$4.57} & \underline{54.71{\scriptsize$\pm$11.68}} & 54.86{\scriptsize$\pm$11.60} & \underline{54.54{\scriptsize$\pm$11.91}} & 55.06{\scriptsize$\pm$11.40} & 11.08{\scriptsize$\pm$6.63} & 22 \\
SEAL(UNI2) & 53.96{\scriptsize$\pm$12.60} & 44.27{\scriptsize$\pm$9.54} & 34.55{\scriptsize$\pm$4.63} & 53.97{\scriptsize$\pm$12.05} & 54.61{\scriptsize$\pm$11.80} & 53.92{\scriptsize$\pm$12.33} & 54.74{\scriptsize$\pm$11.59} & 10.69{\scriptsize$\pm$6.85} & 22 \\
\bottomrule
\end{tabular*}
\end{table}

\begin{table}[H]
\centering
\setlength{\abovecaptionskip}{4pt}
\setlength{\belowcaptionskip}{2pt}
\scriptsize
\caption{\CPBISResultCaption{25}{10}{Macro-AUROC}{AUROC}{fine}}
\label{tab:is_s25_10x_auroc_fine}

\fontsize{7}{8.6}\selectfont
\setlength{\tabcolsep}{1.5pt}
\renewcommand{\arraystretch}{1.0}

\begin{tabular*}{\textwidth}{@{\extracolsep{\fill}}lrrrrrrrrr@{}}
\CPBISResultHeader{AUROC}
CONCH v1.5 & \textbf{73.37{\scriptsize$\pm$11.73}} & 64.28{\scriptsize$\pm$7.47} & 50.05{\scriptsize$\pm$0.19} & 72.65{\scriptsize$\pm$11.71} & \textbf{73.13{\scriptsize$\pm$11.80}} & 72.63{\scriptsize$\pm$12.12} & \textbf{72.95{\scriptsize$\pm$11.78}} & \textbf{9.38{\scriptsize$\pm$5.98}} & 29 \\
CONCH & 73.17{\scriptsize$\pm$11.37} & 64.76{\scriptsize$\pm$7.21} & 50.01{\scriptsize$\pm$0.15} & 72.66{\scriptsize$\pm$11.35} & \underline{73.02{\scriptsize$\pm$11.44}} & 72.47{\scriptsize$\pm$11.73} & 72.73{\scriptsize$\pm$11.50} & \underline{8.75{\scriptsize$\pm$5.86}} & 24 \\
Virchow2 & \underline{73.32{\scriptsize$\pm$11.19}} & 65.05{\scriptsize$\pm$7.17} & 50.05{\scriptsize$\pm$0.24} & 72.47{\scriptsize$\pm$11.51} & 72.60{\scriptsize$\pm$11.46} & 72.55{\scriptsize$\pm$11.69} & 72.54{\scriptsize$\pm$11.45} & 8.63{\scriptsize$\pm$5.67} & 24 \\
Kaiko ViT-S/8 & 71.91{\scriptsize$\pm$11.02} & 63.71{\scriptsize$\pm$7.02} & 49.98{\scriptsize$\pm$0.18} & 72.03{\scriptsize$\pm$11.03} & 71.96{\scriptsize$\pm$10.99} & 71.66{\scriptsize$\pm$11.22} & 71.73{\scriptsize$\pm$11.07} & 8.62{\scriptsize$\pm$5.72} & 20 \\
GPFM & 72.92{\scriptsize$\pm$10.99} & 64.48{\scriptsize$\pm$7.30} & 50.03{\scriptsize$\pm$0.20} & 72.47{\scriptsize$\pm$11.25} & 72.61{\scriptsize$\pm$11.12} & 72.29{\scriptsize$\pm$11.47} & 72.50{\scriptsize$\pm$11.27} & 8.56{\scriptsize$\pm$5.37} & 20 \\
Kaiko ViT-B/8 & 72.60{\scriptsize$\pm$11.43} & 64.49{\scriptsize$\pm$7.13} & \underline{50.10{\scriptsize$\pm$0.33}} & 72.42{\scriptsize$\pm$11.56} & 72.40{\scriptsize$\pm$11.43} & 72.25{\scriptsize$\pm$11.83} & 72.28{\scriptsize$\pm$11.49} & 8.52{\scriptsize$\pm$5.99} & 20 \\
\bottomrule
\end{tabular*}
\end{table}

\begin{table}[H]
\centering
\setlength{\abovecaptionskip}{4pt}
\setlength{\belowcaptionskip}{2pt}
\scriptsize
\caption{\CPBISResultCaption{25}{10}{Macro-AUROC}{AUROC}{coarse}}
\label{tab:is_s25_10x_auroc_coarse}

\fontsize{7}{8.6}\selectfont
\setlength{\tabcolsep}{1.5pt}
\renewcommand{\arraystretch}{1.0}

\begin{tabular*}{\textwidth}{@{\extracolsep{\fill}}lrrrrrrrrr@{}}
\CPBISResultHeader{AUROC}
CONCH & 76.14{\scriptsize$\pm$11.65} & 65.80{\scriptsize$\pm$8.97} & 49.85{\scriptsize$\pm$0.21} & 75.85{\scriptsize$\pm$11.74} & \underline{76.57{\scriptsize$\pm$11.36}} & 75.36{\scriptsize$\pm$12.34} & 76.24{\scriptsize$\pm$11.66} & \textbf{11.17{\scriptsize$\pm$6.25}} & 28 \\
CONCH v1.5 & 75.99{\scriptsize$\pm$12.60} & 66.29{\scriptsize$\pm$8.91} & 50.23{\scriptsize$\pm$0.67} & 75.86{\scriptsize$\pm$12.11} & 76.55{\scriptsize$\pm$11.79} & 75.38{\scriptsize$\pm$12.78} & \textbf{77.19{\scriptsize$\pm$11.29}} & \underline{11.06{\scriptsize$\pm$6.38}} & 28 \\
H-Optimus-1 & 75.90{\scriptsize$\pm$11.81} & 66.53{\scriptsize$\pm$8.87} & 50.38{\scriptsize$\pm$0.78} & 75.36{\scriptsize$\pm$12.03} & 76.27{\scriptsize$\pm$11.50} & 75.20{\scriptsize$\pm$12.42} & 76.08{\scriptsize$\pm$11.65} & 10.46{\scriptsize$\pm$6.28} & 22 \\
Kaiko ViT-S/8 & 75.38{\scriptsize$\pm$11.31} & 65.32{\scriptsize$\pm$8.63} & \underline{50.53{\scriptsize$\pm$1.19}} & 75.56{\scriptsize$\pm$11.31} & 76.06{\scriptsize$\pm$10.68} & 75.11{\scriptsize$\pm$11.68} & 75.63{\scriptsize$\pm$11.15} & 10.42{\scriptsize$\pm$5.59} & 18 \\
SEAL(UNI2) & \textbf{76.86{\scriptsize$\pm$11.03}} & 67.07{\scriptsize$\pm$8.91} & 50.24{\scriptsize$\pm$0.62} & \textbf{76.68{\scriptsize$\pm$11.33}} & 76.46{\scriptsize$\pm$11.29} & \underline{76.50{\scriptsize$\pm$11.73}} & 76.52{\scriptsize$\pm$11.38} & 10.20{\scriptsize$\pm$5.30} & 18 \\
H-Optimus-0 & 76.09{\scriptsize$\pm$11.54} & 66.42{\scriptsize$\pm$8.89} & 50.44{\scriptsize$\pm$0.82} & 75.60{\scriptsize$\pm$11.70} & 75.74{\scriptsize$\pm$11.72} & 75.58{\scriptsize$\pm$12.02} & 75.78{\scriptsize$\pm$11.65} & 10.48{\scriptsize$\pm$5.70} & 17 \\
\bottomrule
\end{tabular*}
\end{table}

\begin{table}[H]
\centering
\setlength{\abovecaptionskip}{4pt}
\setlength{\belowcaptionskip}{2pt}
\scriptsize
\caption{\CPBISResultCaption{25}{10}{Macro-AUPRC}{AUPRC}{fine}}
\label{tab:is_s25_10x_auprc_fine}

\fontsize{7}{8.6}\selectfont
\setlength{\tabcolsep}{1.5pt}
\renewcommand{\arraystretch}{1.0}

\begin{tabular*}{\textwidth}{@{\extracolsep{\fill}}lrrrrrrrrr@{}}
\CPBISResultHeader{AUPRC}
CONCH v1.5 & \textbf{36.12{\scriptsize$\pm$10.40}} & 26.84{\scriptsize$\pm$7.05} & 18.94{\scriptsize$\pm$5.08} & 35.38{\scriptsize$\pm$10.10} & \textbf{35.84{\scriptsize$\pm$10.32}} & \textbf{35.41{\scriptsize$\pm$10.48}} & \textbf{35.68{\scriptsize$\pm$10.29}} & \textbf{9.50{\scriptsize$\pm$6.06}} & 29 \\
Virchow2 & 35.88{\scriptsize$\pm$9.87} & 27.29{\scriptsize$\pm$6.98} & 18.95{\scriptsize$\pm$5.09} & 35.12{\scriptsize$\pm$9.76} & 35.17{\scriptsize$\pm$9.68} & 35.12{\scriptsize$\pm$9.81} & 35.09{\scriptsize$\pm$9.63} & \underline{8.81{\scriptsize$\pm$5.56}} & 25 \\
CONCH & 35.66{\scriptsize$\pm$10.00} & 27.15{\scriptsize$\pm$7.00} & 18.93{\scriptsize$\pm$5.11} & 35.16{\scriptsize$\pm$9.80} & \underline{35.54{\scriptsize$\pm$9.96}} & 35.11{\scriptsize$\pm$10.09} & 35.30{\scriptsize$\pm$9.92} & 8.79{\scriptsize$\pm$5.85} & 25 \\
SEAL(UNI2) & \underline{35.93{\scriptsize$\pm$10.03}} & \underline{27.46{\scriptsize$\pm$7.21}} & 18.93{\scriptsize$\pm$5.07} & \underline{35.53{\scriptsize$\pm$9.98}} & 35.37{\scriptsize$\pm$9.83} & 35.31{\scriptsize$\pm$10.13} & 35.28{\scriptsize$\pm$9.81} & 8.69{\scriptsize$\pm$5.34} & 24 \\
UNI2 & 35.91{\scriptsize$\pm$10.02} & \textbf{27.47{\scriptsize$\pm$7.22}} & 18.94{\scriptsize$\pm$5.08} & \textbf{35.57{\scriptsize$\pm$10.00}} & 35.39{\scriptsize$\pm$9.82} & \underline{35.35{\scriptsize$\pm$10.12}} & \underline{35.35{\scriptsize$\pm$9.80}} & 8.68{\scriptsize$\pm$5.36} & 24 \\
H-Optimus-1 & 35.53{\scriptsize$\pm$9.91} & 27.13{\scriptsize$\pm$7.09} & 18.93{\scriptsize$\pm$5.05} & 34.79{\scriptsize$\pm$9.83} & 34.97{\scriptsize$\pm$9.71} & 34.58{\scriptsize$\pm$9.98} & 34.80{\scriptsize$\pm$9.71} & 8.53{\scriptsize$\pm$5.09} & 23 \\
\bottomrule
\end{tabular*}
\end{table}

\begin{table}[H]
\centering
\setlength{\abovecaptionskip}{4pt}
\setlength{\belowcaptionskip}{2pt}
\scriptsize
\caption{\CPBISResultCaption{25}{10}{Macro-AUPRC}{AUPRC}{coarse}}
\label{tab:is_s25_10x_auprc_coarse}

\fontsize{7}{8.6}\selectfont
\setlength{\tabcolsep}{1.5pt}
\renewcommand{\arraystretch}{1.0}

\begin{tabular*}{\textwidth}{@{\extracolsep{\fill}}lrrrrrrrrr@{}}
\CPBISResultHeader{AUPRC}
CONCH v1.5 & \textbf{58.49{\scriptsize$\pm$19.32}} & 47.60{\scriptsize$\pm$16.17} & 35.96{\scriptsize$\pm$14.53} & 57.85{\scriptsize$\pm$19.06} & \textbf{58.37{\scriptsize$\pm$19.15}} & 57.59{\scriptsize$\pm$19.39} & \textbf{58.23{\scriptsize$\pm$19.15}} & \textbf{12.22{\scriptsize$\pm$7.51}} & 29 \\
CONCH & 58.09{\scriptsize$\pm$18.98} & 47.50{\scriptsize$\pm$16.04} & 35.95{\scriptsize$\pm$14.53} & 57.75{\scriptsize$\pm$18.91} & \underline{58.11{\scriptsize$\pm$18.88}} & 57.32{\scriptsize$\pm$19.21} & \underline{57.96{\scriptsize$\pm$18.96}} & \underline{11.87{\scriptsize$\pm$6.91}} & 28 \\
H-Optimus-1 & 58.02{\scriptsize$\pm$18.80} & 47.83{\scriptsize$\pm$16.41} & 35.96{\scriptsize$\pm$14.53} & 57.29{\scriptsize$\pm$18.82} & 57.71{\scriptsize$\pm$18.86} & 57.05{\scriptsize$\pm$19.09} & 57.49{\scriptsize$\pm$18.87} & 11.46{\scriptsize$\pm$6.70} & 22 \\
Virchow2 & 58.20{\scriptsize$\pm$18.81} & 48.12{\scriptsize$\pm$16.22} & 35.95{\scriptsize$\pm$14.53} & 57.33{\scriptsize$\pm$18.69} & 57.89{\scriptsize$\pm$18.79} & 57.32{\scriptsize$\pm$19.03} & 57.46{\scriptsize$\pm$18.73} & 11.39{\scriptsize$\pm$6.57} & 21 \\
H-Optimus-0 & 58.00{\scriptsize$\pm$18.78} & 47.80{\scriptsize$\pm$16.29} & 35.98{\scriptsize$\pm$14.56} & 57.26{\scriptsize$\pm$18.78} & 57.63{\scriptsize$\pm$18.73} & 56.94{\scriptsize$\pm$19.00} & 57.41{\scriptsize$\pm$18.78} & 11.39{\scriptsize$\pm$6.52} & 21 \\
SEAL(UNI2) & \underline{58.24{\scriptsize$\pm$18.94}} & 48.17{\scriptsize$\pm$16.42} & 35.95{\scriptsize$\pm$14.51} & \textbf{57.87{\scriptsize$\pm$18.94}} & 57.96{\scriptsize$\pm$18.78} & \underline{57.66{\scriptsize$\pm$19.20}} & 57.83{\scriptsize$\pm$18.88} & 11.34{\scriptsize$\pm$6.57} & 21 \\
\bottomrule
\end{tabular*}
\end{table}



\begin{table}[H]
\centering
\setlength{\abovecaptionskip}{4pt}
\setlength{\belowcaptionskip}{2pt}
\scriptsize
\caption{\CPBISResultCaption{20}{40}{Macro-F1}{F1}{fine}}
\label{tab:is_s20_40x_f1_fine}

\fontsize{7}{8.6}\selectfont
\setlength{\tabcolsep}{1.5pt}
\renewcommand{\arraystretch}{1.0}

\begin{tabular*}{\textwidth}{@{\extracolsep{\fill}}lrrrrrrrrr@{}}
\CPBISResultHeader{F1}
UNI2 & \textbf{39.84{\scriptsize$\pm$11.15}} & 30.56{\scriptsize$\pm$7.31} & 15.62{\scriptsize$\pm$6.34} & \underline{40.31{\scriptsize$\pm$11.06}} & \underline{39.69{\scriptsize$\pm$10.87}} & \textbf{40.45{\scriptsize$\pm$11.02}} & \textbf{40.23{\scriptsize$\pm$10.89}} & \textbf{9.20{\scriptsize$\pm$5.48}} & 27 \\
SEAL(UNI2) & \underline{39.82{\scriptsize$\pm$11.16}} & \textbf{30.66{\scriptsize$\pm$7.36}} & 15.68{\scriptsize$\pm$6.27} & 40.30{\scriptsize$\pm$11.04} & 39.57{\scriptsize$\pm$10.87} & \underline{40.43{\scriptsize$\pm$11.04}} & 40.19{\scriptsize$\pm$10.77} & \underline{9.14{\scriptsize$\pm$5.56}} & 27 \\
H-Optimus-1 & 39.75{\scriptsize$\pm$11.09} & \underline{30.57{\scriptsize$\pm$6.76}} & 15.70{\scriptsize$\pm$6.16} & \textbf{40.31{\scriptsize$\pm$10.96}} & \textbf{40.02{\scriptsize$\pm$10.94}} & 40.31{\scriptsize$\pm$10.95} & \underline{40.20{\scriptsize$\pm$10.74}} & 9.08{\scriptsize$\pm$5.81} & 27 \\
Virchow2 & 38.88{\scriptsize$\pm$10.65} & 29.97{\scriptsize$\pm$6.76} & 15.94{\scriptsize$\pm$6.07} & 40.01{\scriptsize$\pm$10.45} & 39.29{\scriptsize$\pm$10.28} & 40.18{\scriptsize$\pm$10.52} & 39.82{\scriptsize$\pm$10.37} & 8.80{\scriptsize$\pm$5.30} & 26 \\
H0mini & 38.69{\scriptsize$\pm$10.78} & 30.04{\scriptsize$\pm$7.05} & 15.81{\scriptsize$\pm$6.15} & 39.26{\scriptsize$\pm$10.59} & 38.89{\scriptsize$\pm$10.61} & 39.42{\scriptsize$\pm$10.72} & 39.47{\scriptsize$\pm$10.65} & 8.57{\scriptsize$\pm$5.24} & 24 \\
H-Optimus-0 & 39.01{\scriptsize$\pm$10.95} & 30.48{\scriptsize$\pm$7.09} & 15.83{\scriptsize$\pm$5.98} & 39.80{\scriptsize$\pm$10.85} & 39.44{\scriptsize$\pm$10.64} & 39.70{\scriptsize$\pm$10.82} & 39.66{\scriptsize$\pm$10.62} & 8.46{\scriptsize$\pm$5.57} & 24 \\
\bottomrule
\end{tabular*}
\end{table}

\begin{table}[H]
\centering
\setlength{\abovecaptionskip}{4pt}
\setlength{\belowcaptionskip}{2pt}
\scriptsize
\caption{\CPBISResultCaption{20}{40}{Macro-F1}{F1}{coarse}}
\label{tab:is_s20_40x_f1_coarse}

\fontsize{7}{8.6}\selectfont
\setlength{\tabcolsep}{1.5pt}
\renewcommand{\arraystretch}{1.0}

\begin{tabular*}{\textwidth}{@{\extracolsep{\fill}}lrrrrrrrrr@{}}
\CPBISResultHeader{F1}
H-Optimus-1 & \textbf{60.82{\scriptsize$\pm$12.20}} & \textbf{53.00{\scriptsize$\pm$10.01}} & 35.33{\scriptsize$\pm$5.34} & \textbf{61.29{\scriptsize$\pm$11.77}} & \textbf{61.32{\scriptsize$\pm$11.46}} & \textbf{61.27{\scriptsize$\pm$11.67}} & \textbf{61.67{\scriptsize$\pm$11.30}} & \textbf{8.56{\scriptsize$\pm$4.20}} & 27 \\
UNI2 & \underline{60.45{\scriptsize$\pm$12.74}} & 52.90{\scriptsize$\pm$10.37} & 34.87{\scriptsize$\pm$5.24} & 60.93{\scriptsize$\pm$12.30} & 60.57{\scriptsize$\pm$12.07} & \underline{61.18{\scriptsize$\pm$12.20}} & 61.29{\scriptsize$\pm$11.88} & 8.40{\scriptsize$\pm$4.01} & 27 \\
SEAL(UNI2) & 60.45{\scriptsize$\pm$12.74} & \underline{52.95{\scriptsize$\pm$10.36}} & 35.15{\scriptsize$\pm$5.33} & \underline{60.98{\scriptsize$\pm$12.27}} & 60.50{\scriptsize$\pm$11.90} & 61.16{\scriptsize$\pm$12.16} & \underline{61.31{\scriptsize$\pm$11.84}} & 8.38{\scriptsize$\pm$4.03} & 27 \\
H-Optimus-0 & 60.14{\scriptsize$\pm$12.32} & 52.40{\scriptsize$\pm$9.98} & 35.44{\scriptsize$\pm$5.12} & 60.79{\scriptsize$\pm$11.59} & \underline{60.69{\scriptsize$\pm$11.37}} & 60.87{\scriptsize$\pm$11.66} & 61.12{\scriptsize$\pm$11.23} & 8.26{\scriptsize$\pm$3.88} & 24 \\
Virchow2 & 59.74{\scriptsize$\pm$12.66} & 51.06{\scriptsize$\pm$10.75} & 35.36{\scriptsize$\pm$5.31} & 60.74{\scriptsize$\pm$11.80} & 60.68{\scriptsize$\pm$11.40} & 60.78{\scriptsize$\pm$11.61} & 61.01{\scriptsize$\pm$11.37} & \underline{8.54{\scriptsize$\pm$3.58}} & 23 \\
Virchow & 59.70{\scriptsize$\pm$12.18} & 52.09{\scriptsize$\pm$9.89} & \underline{35.71{\scriptsize$\pm$5.49}} & 60.55{\scriptsize$\pm$11.61} & 60.61{\scriptsize$\pm$11.36} & 60.76{\scriptsize$\pm$11.41} & 60.74{\scriptsize$\pm$11.20} & 8.29{\scriptsize$\pm$4.12} & 23 \\
\bottomrule
\end{tabular*}
\end{table}

\begin{table}[H]
\centering
\setlength{\abovecaptionskip}{4pt}
\setlength{\belowcaptionskip}{2pt}
\scriptsize
\caption{\CPBISResultCaption{20}{40}{Macro-AUROC}{AUROC}{fine}}
\label{tab:is_s20_40x_auroc_fine}

\fontsize{7}{8.6}\selectfont
\setlength{\tabcolsep}{1.5pt}
\renewcommand{\arraystretch}{1.0}


\end{table}

\begin{table}[H]
\centering
\setlength{\abovecaptionskip}{4pt}
\setlength{\belowcaptionskip}{2pt}
\scriptsize
\caption{\CPBISResultCaption{20}{40}{Macro-AUROC}{AUROC}{coarse}}
\label{tab:is_s20_40x_auroc_coarse}

\fontsize{7}{8.6}\selectfont
\setlength{\tabcolsep}{1.5pt}
\renewcommand{\arraystretch}{1.0}


\end{table}

\begin{table}[H]
\centering
\setlength{\abovecaptionskip}{4pt}
\setlength{\belowcaptionskip}{2pt}
\scriptsize
\caption{\CPBISResultCaption{20}{40}{Macro-AUPRC}{AUPRC}{fine}}
\label{tab:is_s20_40x_auprc_fine}

\fontsize{7}{8.6}\selectfont
\setlength{\tabcolsep}{1.5pt}
\renewcommand{\arraystretch}{1.0}


\end{table}

\begin{table}[H]
\centering
\setlength{\abovecaptionskip}{4pt}
\setlength{\belowcaptionskip}{2pt}
\scriptsize
\caption{\CPBISResultCaption{20}{40}{Macro-AUPRC}{AUPRC}{coarse}}
\label{tab:is_s20_40x_auprc_coarse}

\fontsize{7}{8.6}\selectfont
\setlength{\tabcolsep}{1.5pt}
\renewcommand{\arraystretch}{1.0}


\end{table}


\begin{table}[H]
\centering
\setlength{\abovecaptionskip}{4pt}
\setlength{\belowcaptionskip}{2pt}
\scriptsize
\caption{\CPBISResultCaption{20}{20}{Macro-F1}{F1}{fine}}
\label{tab:is_s20_20x_f1_fine}

\fontsize{7}{8.6}\selectfont
\setlength{\tabcolsep}{1.5pt}
\renewcommand{\arraystretch}{1.0}


\end{table}

\begin{table}[H]
\centering
\setlength{\abovecaptionskip}{4pt}
\setlength{\belowcaptionskip}{2pt}
\scriptsize
\caption{\CPBISResultCaption{20}{20}{Macro-F1}{F1}{coarse}}
\label{tab:is_s20_20x_f1_coarse}

\fontsize{7}{8.6}\selectfont
\setlength{\tabcolsep}{1.5pt}
\renewcommand{\arraystretch}{1.0}


\end{table}

\begin{table}[H]
\centering
\setlength{\abovecaptionskip}{4pt}
\setlength{\belowcaptionskip}{2pt}
\scriptsize
\caption{\CPBISResultCaption{20}{20}{Macro-AUROC}{AUROC}{fine}}
\label{tab:is_s20_20x_auroc_fine}

\fontsize{7}{8.6}\selectfont
\setlength{\tabcolsep}{1.5pt}
\renewcommand{\arraystretch}{1.0}


\end{table}

\begin{table}[H]
\centering
\setlength{\abovecaptionskip}{4pt}
\setlength{\belowcaptionskip}{2pt}
\scriptsize
\caption{\CPBISResultCaption{20}{20}{Macro-AUROC}{AUROC}{coarse}}
\label{tab:is_s20_20x_auroc_coarse}

\fontsize{7}{8.6}\selectfont
\setlength{\tabcolsep}{1.5pt}
\renewcommand{\arraystretch}{1.0}


\end{table}

\begin{table}[H]
\centering
\setlength{\abovecaptionskip}{4pt}
\setlength{\belowcaptionskip}{2pt}
\scriptsize
\caption{\CPBISResultCaption{20}{20}{Macro-AUPRC}{AUPRC}{fine}}
\label{tab:is_s20_20x_auprc_fine}

\fontsize{7}{8.6}\selectfont
\setlength{\tabcolsep}{1.5pt}
\renewcommand{\arraystretch}{1.0}


\end{table}

\begin{table}[H]
\centering
\setlength{\abovecaptionskip}{4pt}
\setlength{\belowcaptionskip}{2pt}
\scriptsize
\caption{\CPBISResultCaption{20}{20}{Macro-AUPRC}{AUPRC}{coarse}}
\label{tab:is_s20_20x_auprc_coarse}

\fontsize{7}{8.6}\selectfont
\setlength{\tabcolsep}{1.5pt}
\renewcommand{\arraystretch}{1.0}


\end{table}


\begin{table}[H]
\centering
\setlength{\abovecaptionskip}{4pt}
\setlength{\belowcaptionskip}{2pt}
\scriptsize
\caption{\CPBISResultCaption{20}{10}{Macro-F1}{F1}{fine}}
\label{tab:is_s20_10x_f1_fine}

\fontsize{7}{8.6}\selectfont
\setlength{\tabcolsep}{1.5pt}
\renewcommand{\arraystretch}{1.0}

\begin{tabular*}{\textwidth}{@{\extracolsep{\fill}}lrrrrrrrrr@{}}
\CPBISResultHeader{F1}
CONCH v1.5 & \textbf{35.81{\scriptsize$\pm$9.52}} & 24.26{\scriptsize$\pm$6.28} & 15.78{\scriptsize$\pm$6.07} & \textbf{35.89{\scriptsize$\pm$9.07}} & \textbf{36.55{\scriptsize$\pm$9.27}} & \textbf{36.42{\scriptsize$\pm$9.37}} & \textbf{36.61{\scriptsize$\pm$9.18}} & \textbf{11.57{\scriptsize$\pm$6.24}} & 28 \\
Virchow2 & \underline{35.48{\scriptsize$\pm$8.74}} & 24.58{\scriptsize$\pm$6.26} & 15.71{\scriptsize$\pm$6.05} & \underline{35.86{\scriptsize$\pm$8.43}} & 35.98{\scriptsize$\pm$8.40} & \underline{36.12{\scriptsize$\pm$8.36}} & \underline{36.12{\scriptsize$\pm$8.33}} & \underline{11.21{\scriptsize$\pm$5.49}} & 27 \\
CONCH & 35.31{\scriptsize$\pm$9.12} & 24.29{\scriptsize$\pm$6.09} & 15.58{\scriptsize$\pm$6.14} & 35.62{\scriptsize$\pm$8.70} & \underline{36.05{\scriptsize$\pm$8.90}} & 35.81{\scriptsize$\pm$8.92} & 36.06{\scriptsize$\pm$8.84} & 11.09{\scriptsize$\pm$5.69} & 23 \\
UNI2 & 35.31{\scriptsize$\pm$8.86} & 24.52{\scriptsize$\pm$6.18} & 15.23{\scriptsize$\pm$6.37} & 35.60{\scriptsize$\pm$8.49} & 35.83{\scriptsize$\pm$8.75} & 35.67{\scriptsize$\pm$8.69} & 35.91{\scriptsize$\pm$8.60} & 10.99{\scriptsize$\pm$5.45} & 23 \\
SEAL(UNI2) & 35.33{\scriptsize$\pm$8.89} & 24.53{\scriptsize$\pm$6.26} & 15.40{\scriptsize$\pm$6.40} & 35.56{\scriptsize$\pm$8.48} & 35.84{\scriptsize$\pm$8.74} & 35.65{\scriptsize$\pm$8.71} & 35.95{\scriptsize$\pm$8.62} & 10.96{\scriptsize$\pm$5.45} & 23 \\
H-Optimus-1 & 35.23{\scriptsize$\pm$8.86} & 24.48{\scriptsize$\pm$5.97} & 15.40{\scriptsize$\pm$6.16} & 35.43{\scriptsize$\pm$8.56} & 35.70{\scriptsize$\pm$8.54} & 35.42{\scriptsize$\pm$8.74} & 35.58{\scriptsize$\pm$8.46} & 10.93{\scriptsize$\pm$5.34} & 23 \\
\bottomrule
\end{tabular*}
\end{table}

\begin{table}[H]
\centering
\setlength{\abovecaptionskip}{4pt}
\setlength{\belowcaptionskip}{2pt}
\scriptsize
\caption{\CPBISResultCaption{20}{10}{Macro-F1}{F1}{coarse}}
\label{tab:is_s20_10x_f1_coarse}

\fontsize{7}{8.6}\selectfont
\setlength{\tabcolsep}{1.5pt}
\renewcommand{\arraystretch}{1.0}


\end{table}

\begin{table}[H]
\centering
\setlength{\abovecaptionskip}{4pt}
\setlength{\belowcaptionskip}{2pt}
\scriptsize
\caption{\CPBISResultCaption{20}{10}{Macro-AUROC}{AUROC}{fine}}
\label{tab:is_s20_10x_auroc_fine}

\fontsize{7}{8.6}\selectfont
\setlength{\tabcolsep}{1.5pt}
\renewcommand{\arraystretch}{1.0}


\end{table}

\begin{table}[H]
\centering
\setlength{\abovecaptionskip}{4pt}
\setlength{\belowcaptionskip}{2pt}
\scriptsize
\caption{\CPBISResultCaption{20}{10}{Macro-AUROC}{AUROC}{coarse}}
\label{tab:is_s20_10x_auroc_coarse}

\fontsize{7}{8.6}\selectfont
\setlength{\tabcolsep}{1.5pt}
\renewcommand{\arraystretch}{1.0}


\end{table}

\begin{table}[H]
\centering
\setlength{\abovecaptionskip}{4pt}
\setlength{\belowcaptionskip}{2pt}
\scriptsize
\caption{\CPBISResultCaption{20}{10}{Macro-AUPRC}{AUPRC}{fine}}
\label{tab:is_s20_10x_auprc_fine}

\fontsize{7}{8.6}\selectfont
\setlength{\tabcolsep}{1.5pt}
\renewcommand{\arraystretch}{1.0}


\end{table}

\begin{table}[H]
\centering
\setlength{\abovecaptionskip}{4pt}
\setlength{\belowcaptionskip}{2pt}
\scriptsize
\caption{\CPBISResultCaption{20}{10}{Macro-AUPRC}{AUPRC}{coarse}}
\label{tab:is_s20_10x_auprc_coarse}

\fontsize{7}{8.6}\selectfont
\setlength{\tabcolsep}{1.5pt}
\renewcommand{\arraystretch}{1.0}


\end{table}



\begin{table}[H]
\centering
\setlength{\abovecaptionskip}{4pt}
\setlength{\belowcaptionskip}{2pt}
\scriptsize
\caption{\CPBISResultCaption{15}{40}{Macro-F1}{F1}{fine}}
\label{tab:is_s15_40x_f1_fine}

\fontsize{7}{8.6}\selectfont
\setlength{\tabcolsep}{1.5pt}
\renewcommand{\arraystretch}{1.0}


\end{table}

\begin{table}[H]
\centering
\setlength{\abovecaptionskip}{4pt}
\setlength{\belowcaptionskip}{2pt}
\scriptsize
\caption{\CPBISResultCaption{15}{40}{Macro-F1}{F1}{coarse}}
\label{tab:is_s15_40x_f1_coarse}

\fontsize{7}{8.6}\selectfont
\setlength{\tabcolsep}{1.5pt}
\renewcommand{\arraystretch}{1.0}


\end{table}

\begin{table}[H]
\centering
\setlength{\abovecaptionskip}{4pt}
\setlength{\belowcaptionskip}{2pt}
\scriptsize
\caption{\CPBISResultCaption{15}{40}{Macro-AUROC}{AUROC}{fine}}
\label{tab:is_s15_40x_auroc_fine}

\fontsize{7}{8.6}\selectfont
\setlength{\tabcolsep}{1.5pt}
\renewcommand{\arraystretch}{1.0}


\end{table}

\begin{table}[H]
\centering
\setlength{\abovecaptionskip}{4pt}
\setlength{\belowcaptionskip}{2pt}
\scriptsize
\caption{\CPBISResultCaption{15}{40}{Macro-AUROC}{AUROC}{coarse}}
\label{tab:is_s15_40x_auroc_coarse}

\fontsize{7}{8.6}\selectfont
\setlength{\tabcolsep}{1.5pt}
\renewcommand{\arraystretch}{1.0}


\end{table}

\begin{table}[H]
\centering
\setlength{\abovecaptionskip}{4pt}
\setlength{\belowcaptionskip}{2pt}
\scriptsize
\caption{\CPBISResultCaption{15}{40}{Macro-AUPRC}{AUPRC}{fine}}
\label{tab:is_s15_40x_auprc_fine}

\fontsize{7}{8.6}\selectfont
\setlength{\tabcolsep}{1.5pt}
\renewcommand{\arraystretch}{1.0}


\end{table}

\begin{table}[H]
\centering
\setlength{\abovecaptionskip}{4pt}
\setlength{\belowcaptionskip}{2pt}
\scriptsize
\caption{\CPBISResultCaption{15}{40}{Macro-AUPRC}{AUPRC}{coarse}}
\label{tab:is_s15_40x_auprc_coarse}

\fontsize{7}{8.6}\selectfont
\setlength{\tabcolsep}{1.5pt}
\renewcommand{\arraystretch}{1.0}


\end{table}


\begin{table}[H]
\centering
\setlength{\abovecaptionskip}{4pt}
\setlength{\belowcaptionskip}{2pt}
\scriptsize
\caption{\CPBISResultCaption{15}{20}{Macro-F1}{F1}{fine}}
\label{tab:is_s15_20x_f1_fine}

\fontsize{7}{8.6}\selectfont
\setlength{\tabcolsep}{1.5pt}
\renewcommand{\arraystretch}{1.0}


\end{table}

\begin{table}[H]
\centering
\setlength{\abovecaptionskip}{4pt}
\setlength{\belowcaptionskip}{2pt}
\scriptsize
\caption{\CPBISResultCaption{15}{20}{Macro-F1}{F1}{coarse}}
\label{tab:is_s15_20x_f1_coarse}

\fontsize{7}{8.6}\selectfont
\setlength{\tabcolsep}{1.5pt}
\renewcommand{\arraystretch}{1.0}


\end{table}

\begin{table}[H]
\centering
\setlength{\abovecaptionskip}{4pt}
\setlength{\belowcaptionskip}{2pt}
\scriptsize
\caption{\CPBISResultCaption{15}{20}{Macro-AUROC}{AUROC}{fine}}
\label{tab:is_s15_20x_auroc_fine}

\fontsize{7}{8.6}\selectfont
\setlength{\tabcolsep}{1.5pt}
\renewcommand{\arraystretch}{1.0}


\end{table}

\begin{table}[H]
\centering
\setlength{\abovecaptionskip}{4pt}
\setlength{\belowcaptionskip}{2pt}
\scriptsize
\caption{\CPBISResultCaption{15}{20}{Macro-AUROC}{AUROC}{coarse}}
\label{tab:is_s15_20x_auroc_coarse}

\fontsize{7}{8.6}\selectfont
\setlength{\tabcolsep}{1.5pt}
\renewcommand{\arraystretch}{1.0}


\end{table}

\begin{table}[H]
\centering
\setlength{\abovecaptionskip}{4pt}
\setlength{\belowcaptionskip}{2pt}
\scriptsize
\caption{\CPBISResultCaption{15}{20}{Macro-AUPRC}{AUPRC}{fine}}
\label{tab:is_s15_20x_auprc_fine}

\fontsize{7}{8.6}\selectfont
\setlength{\tabcolsep}{1.5pt}
\renewcommand{\arraystretch}{1.0}


\end{table}

\begin{table}[H]
\centering
\setlength{\abovecaptionskip}{4pt}
\setlength{\belowcaptionskip}{2pt}
\scriptsize
\caption{\CPBISResultCaption{15}{20}{Macro-AUPRC}{AUPRC}{coarse}}
\label{tab:is_s15_20x_auprc_coarse}

\fontsize{7}{8.6}\selectfont
\setlength{\tabcolsep}{1.5pt}
\renewcommand{\arraystretch}{1.0}


\end{table}


\begin{table}[H]
\centering
\setlength{\abovecaptionskip}{4pt}
\setlength{\belowcaptionskip}{2pt}
\scriptsize
\caption{\CPBISResultCaption{15}{10}{Macro-F1}{F1}{fine}}
\label{tab:is_s15_10x_f1_fine}

\fontsize{7}{8.6}\selectfont
\setlength{\tabcolsep}{1.5pt}
\renewcommand{\arraystretch}{1.0}


\end{table}

\begin{table}[H]
\centering
\setlength{\abovecaptionskip}{4pt}
\setlength{\belowcaptionskip}{2pt}
\scriptsize
\caption{\CPBISResultCaption{15}{10}{Macro-F1}{F1}{coarse}}
\label{tab:is_s15_10x_f1_coarse}

\fontsize{7}{8.6}\selectfont
\setlength{\tabcolsep}{1.5pt}
\renewcommand{\arraystretch}{1.0}


\end{table}

\begin{table}[H]
\centering
\setlength{\abovecaptionskip}{4pt}
\setlength{\belowcaptionskip}{2pt}
\scriptsize
\caption{\CPBISResultCaption{15}{10}{Macro-AUROC}{AUROC}{fine}}
\label{tab:is_s15_10x_auroc_fine}

\fontsize{7}{8.6}\selectfont
\setlength{\tabcolsep}{1.5pt}
\renewcommand{\arraystretch}{1.0}


\end{table}

\begin{table}[H]
\centering
\setlength{\abovecaptionskip}{4pt}
\setlength{\belowcaptionskip}{2pt}
\scriptsize
\caption{\CPBISResultCaption{15}{10}{Macro-AUROC}{AUROC}{coarse}}
\label{tab:is_s15_10x_auroc_coarse}

\fontsize{7}{8.6}\selectfont
\setlength{\tabcolsep}{1.5pt}
\renewcommand{\arraystretch}{1.0}


\end{table}

\begin{table}[H]
\centering
\setlength{\abovecaptionskip}{4pt}
\setlength{\belowcaptionskip}{2pt}
\scriptsize
\caption{\CPBISResultCaption{15}{10}{Macro-AUPRC}{AUPRC}{fine}}
\label{tab:is_s15_10x_auprc_fine}

\fontsize{7}{8.6}\selectfont
\setlength{\tabcolsep}{1.5pt}
\renewcommand{\arraystretch}{1.0}


\end{table}

\begin{table}[H]
\centering
\setlength{\abovecaptionskip}{4pt}
\setlength{\belowcaptionskip}{2pt}
\scriptsize
\caption{\CPBISResultCaption{15}{10}{Macro-AUPRC}{AUPRC}{coarse}}
\label{tab:is_s15_10x_auprc_coarse}

\fontsize{7}{8.6}\selectfont
\setlength{\tabcolsep}{1.5pt}
\renewcommand{\arraystretch}{1.0}


\end{table}



\begin{table}[H]
\centering
\setlength{\abovecaptionskip}{4pt}
\setlength{\belowcaptionskip}{2pt}
\scriptsize
\caption{\CPBISResultCaption{10}{40}{Macro-F1}{F1}{fine}}
\label{tab:is_s10_40x_f1_fine}

\fontsize{7}{8.6}\selectfont
\setlength{\tabcolsep}{1.5pt}
\renewcommand{\arraystretch}{1.0}


\end{table}

\begin{table}[H]
\centering
\setlength{\abovecaptionskip}{4pt}
\setlength{\belowcaptionskip}{2pt}
\scriptsize
\caption{\CPBISResultCaption{10}{40}{Macro-F1}{F1}{coarse}}
\label{tab:is_s10_40x_f1_coarse}

\fontsize{7}{8.6}\selectfont
\setlength{\tabcolsep}{1.5pt}
\renewcommand{\arraystretch}{1.0}


\end{table}

\begin{table}[H]
\centering
\setlength{\abovecaptionskip}{4pt}
\setlength{\belowcaptionskip}{2pt}
\scriptsize
\caption{\CPBISResultCaption{10}{40}{Macro-AUROC}{AUROC}{fine}}
\label{tab:is_s10_40x_auroc_fine}

\fontsize{7}{8.6}\selectfont
\setlength{\tabcolsep}{1.5pt}
\renewcommand{\arraystretch}{1.0}


\end{table}

\begin{table}[H]
\centering
\setlength{\abovecaptionskip}{4pt}
\setlength{\belowcaptionskip}{2pt}
\scriptsize
\caption{\CPBISResultCaption{10}{40}{Macro-AUROC}{AUROC}{coarse}}
\label{tab:is_s10_40x_auroc_coarse}

\fontsize{7}{8.6}\selectfont
\setlength{\tabcolsep}{1.5pt}
\renewcommand{\arraystretch}{1.0}


\end{table}

\begin{table}[H]
\centering
\setlength{\abovecaptionskip}{4pt}
\setlength{\belowcaptionskip}{2pt}
\scriptsize
\caption{\CPBISResultCaption{10}{40}{Macro-AUPRC}{AUPRC}{fine}}
\label{tab:is_s10_40x_auprc_fine}

\fontsize{7}{8.6}\selectfont
\setlength{\tabcolsep}{1.5pt}
\renewcommand{\arraystretch}{1.0}


\end{table}

\begin{table}[H]
\centering
\setlength{\abovecaptionskip}{4pt}
\setlength{\belowcaptionskip}{2pt}
\scriptsize
\caption{\CPBISResultCaption{10}{40}{Macro-AUPRC}{AUPRC}{coarse}}
\label{tab:is_s10_40x_auprc_coarse}

\fontsize{7}{8.6}\selectfont
\setlength{\tabcolsep}{1.5pt}
\renewcommand{\arraystretch}{1.0}


\end{table}


\begin{table}[H]
\centering
\setlength{\abovecaptionskip}{4pt}
\setlength{\belowcaptionskip}{2pt}
\scriptsize
\caption{\CPBISResultCaption{10}{20}{Macro-F1}{F1}{fine}}
\label{tab:is_s10_20x_f1_fine}

\fontsize{7}{8.6}\selectfont
\setlength{\tabcolsep}{1.5pt}
\renewcommand{\arraystretch}{1.0}


\end{table}

\begin{table}[H]
\centering
\setlength{\abovecaptionskip}{4pt}
\setlength{\belowcaptionskip}{2pt}
\scriptsize
\caption{\CPBISResultCaption{10}{20}{Macro-F1}{F1}{coarse}}
\label{tab:is_s10_20x_f1_coarse}

\fontsize{7}{8.6}\selectfont
\setlength{\tabcolsep}{1.5pt}
\renewcommand{\arraystretch}{1.0}


\end{table}

\begin{table}[H]
\centering
\setlength{\abovecaptionskip}{4pt}
\setlength{\belowcaptionskip}{2pt}
\scriptsize
\caption{\CPBISResultCaption{10}{20}{Macro-AUROC}{AUROC}{fine}}
\label{tab:is_s10_20x_auroc_fine}

\fontsize{7}{8.6}\selectfont
\setlength{\tabcolsep}{1.5pt}
\renewcommand{\arraystretch}{1.0}


\end{table}

\begin{table}[H]
\centering
\setlength{\abovecaptionskip}{4pt}
\setlength{\belowcaptionskip}{2pt}
\scriptsize
\caption{\CPBISResultCaption{10}{20}{Macro-AUROC}{AUROC}{coarse}}
\label{tab:is_s10_20x_auroc_coarse}

\fontsize{7}{8.6}\selectfont
\setlength{\tabcolsep}{1.5pt}
\renewcommand{\arraystretch}{1.0}


\end{table}

\begin{table}[H]
\centering
\setlength{\abovecaptionskip}{4pt}
\setlength{\belowcaptionskip}{2pt}
\scriptsize
\caption{\CPBISResultCaption{10}{20}{Macro-AUPRC}{AUPRC}{fine}}
\label{tab:is_s10_20x_auprc_fine}

\fontsize{7}{8.6}\selectfont
\setlength{\tabcolsep}{1.5pt}
\renewcommand{\arraystretch}{1.0}


\end{table}

\begin{table}[H]
\centering
\setlength{\abovecaptionskip}{4pt}
\setlength{\belowcaptionskip}{2pt}
\scriptsize
\caption{\CPBISResultCaption{10}{20}{Macro-AUPRC}{AUPRC}{coarse}}
\label{tab:is_s10_20x_auprc_coarse}

\fontsize{7}{8.6}\selectfont
\setlength{\tabcolsep}{1.5pt}
\renewcommand{\arraystretch}{1.0}


\end{table}


\begin{table}[H]
\centering
\setlength{\abovecaptionskip}{4pt}
\setlength{\belowcaptionskip}{2pt}
\scriptsize
\caption{\CPBISResultCaption{10}{10}{Macro-F1}{F1}{fine}}
\label{tab:is_s10_10x_f1_fine}

\fontsize{7}{8.6}\selectfont
\setlength{\tabcolsep}{1.5pt}
\renewcommand{\arraystretch}{1.0}


\end{table}

\begin{table}[H]
\centering
\setlength{\abovecaptionskip}{4pt}
\setlength{\belowcaptionskip}{2pt}
\scriptsize
\caption{\CPBISResultCaption{10}{10}{Macro-F1}{F1}{coarse}}
\label{tab:is_s10_10x_f1_coarse}

\fontsize{7}{8.6}\selectfont
\setlength{\tabcolsep}{1.5pt}
\renewcommand{\arraystretch}{1.0}


\end{table}

\begin{table}[H]
\centering
\setlength{\abovecaptionskip}{4pt}
\setlength{\belowcaptionskip}{2pt}
\scriptsize
\caption{\CPBISResultCaption{10}{10}{Macro-AUROC}{AUROC}{fine}}
\label{tab:is_s10_10x_auroc_fine}

\fontsize{7}{8.6}\selectfont
\setlength{\tabcolsep}{1.5pt}
\renewcommand{\arraystretch}{1.0}


\end{table}

\begin{table}[H]
\centering
\setlength{\abovecaptionskip}{4pt}
\setlength{\belowcaptionskip}{2pt}
\scriptsize
\caption{\CPBISResultCaption{10}{10}{Macro-AUROC}{AUROC}{coarse}}
\label{tab:is_s10_10x_auroc_coarse}

\fontsize{7}{8.6}\selectfont
\setlength{\tabcolsep}{1.5pt}
\renewcommand{\arraystretch}{1.0}


\end{table}

\begin{table}[H]
\centering
\setlength{\abovecaptionskip}{4pt}
\setlength{\belowcaptionskip}{2pt}
\scriptsize
\caption{\CPBISResultCaption{10}{10}{Macro-AUPRC}{AUPRC}{fine}}
\label{tab:is_s10_10x_auprc_fine}

\fontsize{7}{8.6}\selectfont
\setlength{\tabcolsep}{1.5pt}
\renewcommand{\arraystretch}{1.0}


\end{table}

\begin{table}[H]
\centering
\setlength{\abovecaptionskip}{4pt}
\setlength{\belowcaptionskip}{2pt}
\scriptsize
\caption{\CPBISResultCaption{10}{10}{Macro-AUPRC}{AUPRC}{coarse}}
\label{tab:is_s10_10x_auprc_coarse}

\fontsize{7}{8.6}\selectfont
\setlength{\tabcolsep}{1.5pt}
\renewcommand{\arraystretch}{1.0}


\end{table}

\endgroup

%
%

\providecommand{\CrossResultsRoot}{supplementary/cross_results}

%

\newcommand{\CPBXResultCaption}[4]{%
Complete \textbf{#1} results at \(\mathbf{#2\times}\) magnification,
\(\mathbf{T_{\mathrm{#4}}}\) taxonomy, \textbf{#3}. See the shared legend above.%
}

\newcommand{\CPBXResultHeader}{%
\toprule
Model & Nuc & Mean & CLS & N+M & N+C & N\(\Vert\)M & N\(\Vert\)C & Wins \\
\midrule
}

\begingroup
\setlength{\intextsep}{6pt}


\noindent
The following tables report the complete cross-section results for all 30 frozen
pathology foundation models under the three transfer protocols. They are organised by
protocol (\textbf{IOCD}, \textbf{MOCV}, \textbf{LOOO}), then by magnification
(\(40\times\), \(20\times\), \(10\times\)), then by metric (Macro-F1, Macro-AUROC,
Macro-AUPRC), and finally by taxonomy (\(T_{\mathrm{fine}}\),
\(T_{\mathrm{coarse}}\)).

\smallskip\noindent
\textbf{Protocols.} \textbf{IOCD} is intra-organ cross-dataset transfer: training on
the remaining sections of an organ and testing on one held-out section, over 23
held-out sections. \textbf{MOCV} is multi-organ cross-validation over five folds
pooled across organs. \textbf{LOOO} is leave-one-organ-out transfer over 11 held-out
organs. Every protocol is run with three seeds.

\smallskip\noindent
\textbf{Taxonomies.} \(T_{\mathrm{fine}}\) is the 9-class H\&E-aligned taxonomy and
\(T_{\mathrm{coarse}}\) its 3-class coarsening. Within an evaluation unit a class is
scored only if it meets the minimum training and test cell counts, so the number of
scored classes can vary by unit.

\smallskip\noindent
\textbf{Readouts (column headings).} Each column is one way of turning a frozen
encoder's output into a per-cell vector.
\textbf{Nuc} samples the dense feature map at the registered nuclear centre.
\textbf{Mean} averages all spatial tokens of the patch.
\textbf{CLS} uses the architecture's native class token.
\textbf{N\(+\)M} and \textbf{N\(+\)C} are additive fusions of the nucleus readout with
the mean-pooled and the class-token readout respectively, whereas
\textbf{N\(\Vert\)M} and \textbf{N\(\Vert\)C} are the corresponding concatenative
fusions. A dash (--) marks a class-token-dependent readout that is structurally
unavailable, which applies to the one model without a native class token.

\smallskip\noindent
\textbf{Aggregation.} Each value is the balanced mean over the evaluation units of its
protocol, with the dispersion reported as the sample standard deviation over those
same units: organ-balanced for IOCD, fold-balanced for MOCV and
held-out-organ-balanced for LOOO. All values are in percentage points.

\smallskip\noindent
\textbf{Significant wins (Wins).} The last column counts how many of the other 29
models a given model significantly exceeds on the \textbf{Nuc} readout, by a two-sided
Wilcoxon signed-rank test on their paired per-unit values over the common
fold \(\times\) seed units, at \(p < 0.05\) and with a positive median difference. The
maximum is therefore 29. The test uses the metric and the taxonomy of the table it
appears in, so no value in a table depends on any other table.

\smallskip\noindent
\textbf{Row order.} Models are sorted by Wins in descending order, with the unrounded
Nuc mean as the tie-break.

\smallskip\noindent
\textbf{Highlighting.} Within each readout column the best value is set in
\textbf{bold} and the second best is \underline{underlined}; higher is better
throughout. The Wins column is not highlighted, because it is the sort key.



\begin{table}[H]
\centering
\setlength{\abovecaptionskip}{4pt}
\setlength{\belowcaptionskip}{2pt}
\scriptsize
\caption{\CPBXResultCaption{IOCD}{40}{Macro-F1}{fine}}
\label{tab:cross_iocd_40x_f1_fine}

\fontsize{7}{8.6}\selectfont
\setlength{\tabcolsep}{1.5pt}
\renewcommand{\arraystretch}{1.0}


\end{table}

\begin{table}[H]
\centering
\setlength{\abovecaptionskip}{4pt}
\setlength{\belowcaptionskip}{2pt}
\scriptsize
\caption{\CPBXResultCaption{IOCD}{40}{Macro-F1}{coarse}}
\label{tab:cross_iocd_40x_f1_coarse}

\fontsize{7}{8.6}\selectfont
\setlength{\tabcolsep}{1.5pt}
\renewcommand{\arraystretch}{1.0}


\end{table}

\begin{table}[H]
\centering
\setlength{\abovecaptionskip}{4pt}
\setlength{\belowcaptionskip}{2pt}
\scriptsize
\caption{\CPBXResultCaption{IOCD}{40}{Macro-AUROC}{fine}}
\label{tab:cross_iocd_40x_auroc_fine}

\fontsize{7}{8.6}\selectfont
\setlength{\tabcolsep}{1.5pt}
\renewcommand{\arraystretch}{1.0}


\end{table}

\begin{table}[H]
\centering
\setlength{\abovecaptionskip}{4pt}
\setlength{\belowcaptionskip}{2pt}
\scriptsize
\caption{\CPBXResultCaption{IOCD}{40}{Macro-AUROC}{coarse}}
\label{tab:cross_iocd_40x_auroc_coarse}

\fontsize{7}{8.6}\selectfont
\setlength{\tabcolsep}{1.5pt}
\renewcommand{\arraystretch}{1.0}


\end{table}

\begin{table}[H]
\centering
\setlength{\abovecaptionskip}{4pt}
\setlength{\belowcaptionskip}{2pt}
\scriptsize
\caption{\CPBXResultCaption{IOCD}{40}{Macro-AUPRC}{fine}}
\label{tab:cross_iocd_40x_auprc_fine}

\fontsize{7}{8.6}\selectfont
\setlength{\tabcolsep}{1.5pt}
\renewcommand{\arraystretch}{1.0}


\end{table}

\begin{table}[H]
\centering
\setlength{\abovecaptionskip}{4pt}
\setlength{\belowcaptionskip}{2pt}
\scriptsize
\caption{\CPBXResultCaption{IOCD}{40}{Macro-AUPRC}{coarse}}
\label{tab:cross_iocd_40x_auprc_coarse}

\fontsize{7}{8.6}\selectfont
\setlength{\tabcolsep}{1.5pt}
\renewcommand{\arraystretch}{1.0}


\end{table}


\begin{table}[H]
\centering
\setlength{\abovecaptionskip}{4pt}
\setlength{\belowcaptionskip}{2pt}
\scriptsize
\caption{\CPBXResultCaption{IOCD}{20}{Macro-F1}{fine}}
\label{tab:cross_iocd_20x_f1_fine}

\fontsize{7}{8.6}\selectfont
\setlength{\tabcolsep}{1.5pt}
\renewcommand{\arraystretch}{1.0}


\end{table}

\begin{table}[H]
\centering
\setlength{\abovecaptionskip}{4pt}
\setlength{\belowcaptionskip}{2pt}
\scriptsize
\caption{\CPBXResultCaption{IOCD}{20}{Macro-F1}{coarse}}
\label{tab:cross_iocd_20x_f1_coarse}

\fontsize{7}{8.6}\selectfont
\setlength{\tabcolsep}{1.5pt}
\renewcommand{\arraystretch}{1.0}


\end{table}

\begin{table}[H]
\centering
\setlength{\abovecaptionskip}{4pt}
\setlength{\belowcaptionskip}{2pt}
\scriptsize
\caption{\CPBXResultCaption{IOCD}{20}{Macro-AUROC}{fine}}
\label{tab:cross_iocd_20x_auroc_fine}

\fontsize{7}{8.6}\selectfont
\setlength{\tabcolsep}{1.5pt}
\renewcommand{\arraystretch}{1.0}


\end{table}

\begin{table}[H]
\centering
\setlength{\abovecaptionskip}{4pt}
\setlength{\belowcaptionskip}{2pt}
\scriptsize
\caption{\CPBXResultCaption{IOCD}{20}{Macro-AUROC}{coarse}}
\label{tab:cross_iocd_20x_auroc_coarse}

\fontsize{7}{8.6}\selectfont
\setlength{\tabcolsep}{1.5pt}
\renewcommand{\arraystretch}{1.0}


\end{table}

\begin{table}[H]
\centering
\setlength{\abovecaptionskip}{4pt}
\setlength{\belowcaptionskip}{2pt}
\scriptsize
\caption{\CPBXResultCaption{IOCD}{20}{Macro-AUPRC}{fine}}
\label{tab:cross_iocd_20x_auprc_fine}

\fontsize{7}{8.6}\selectfont
\setlength{\tabcolsep}{1.5pt}
\renewcommand{\arraystretch}{1.0}


\end{table}

\begin{table}[H]
\centering
\setlength{\abovecaptionskip}{4pt}
\setlength{\belowcaptionskip}{2pt}
\scriptsize
\caption{\CPBXResultCaption{IOCD}{20}{Macro-AUPRC}{coarse}}
\label{tab:cross_iocd_20x_auprc_coarse}

\fontsize{7}{8.6}\selectfont
\setlength{\tabcolsep}{1.5pt}
\renewcommand{\arraystretch}{1.0}


\end{table}


\begin{table}[H]
\centering
\setlength{\abovecaptionskip}{4pt}
\setlength{\belowcaptionskip}{2pt}
\scriptsize
\caption{\CPBXResultCaption{IOCD}{10}{Macro-F1}{fine}}
\label{tab:cross_iocd_10x_f1_fine}

\fontsize{7}{8.6}\selectfont
\setlength{\tabcolsep}{1.5pt}
\renewcommand{\arraystretch}{1.0}


\end{table}

\begin{table}[H]
\centering
\setlength{\abovecaptionskip}{4pt}
\setlength{\belowcaptionskip}{2pt}
\scriptsize
\caption{\CPBXResultCaption{IOCD}{10}{Macro-F1}{coarse}}
\label{tab:cross_iocd_10x_f1_coarse}

\fontsize{7}{8.6}\selectfont
\setlength{\tabcolsep}{1.5pt}
\renewcommand{\arraystretch}{1.0}


\end{table}

\begin{table}[H]
\centering
\setlength{\abovecaptionskip}{4pt}
\setlength{\belowcaptionskip}{2pt}
\scriptsize
\caption{\CPBXResultCaption{IOCD}{10}{Macro-AUROC}{fine}}
\label{tab:cross_iocd_10x_auroc_fine}

\fontsize{7}{8.6}\selectfont
\setlength{\tabcolsep}{1.5pt}
\renewcommand{\arraystretch}{1.0}


\end{table}

\begin{table}[H]
\centering
\setlength{\abovecaptionskip}{4pt}
\setlength{\belowcaptionskip}{2pt}
\scriptsize
\caption{\CPBXResultCaption{IOCD}{10}{Macro-AUROC}{coarse}}
\label{tab:cross_iocd_10x_auroc_coarse}

\fontsize{7}{8.6}\selectfont
\setlength{\tabcolsep}{1.5pt}
\renewcommand{\arraystretch}{1.0}


\end{table}

\begin{table}[H]
\centering
\setlength{\abovecaptionskip}{4pt}
\setlength{\belowcaptionskip}{2pt}
\scriptsize
\caption{\CPBXResultCaption{IOCD}{10}{Macro-AUPRC}{fine}}
\label{tab:cross_iocd_10x_auprc_fine}

\fontsize{7}{8.6}\selectfont
\setlength{\tabcolsep}{1.5pt}
\renewcommand{\arraystretch}{1.0}


\end{table}

\begin{table}[H]
\centering
\setlength{\abovecaptionskip}{4pt}
\setlength{\belowcaptionskip}{2pt}
\scriptsize
\caption{\CPBXResultCaption{IOCD}{10}{Macro-AUPRC}{coarse}}
\label{tab:cross_iocd_10x_auprc_coarse}

\fontsize{7}{8.6}\selectfont
\setlength{\tabcolsep}{1.5pt}
\renewcommand{\arraystretch}{1.0}


\end{table}



\begin{table}[H]
\centering
\setlength{\abovecaptionskip}{4pt}
\setlength{\belowcaptionskip}{2pt}
\scriptsize
\caption{\CPBXResultCaption{MOCV}{40}{Macro-F1}{fine}}
\label{tab:cross_mocv_40x_f1_fine}

\fontsize{7}{8.6}\selectfont
\setlength{\tabcolsep}{1.5pt}
\renewcommand{\arraystretch}{1.0}


\end{table}

\begin{table}[H]
\centering
\setlength{\abovecaptionskip}{4pt}
\setlength{\belowcaptionskip}{2pt}
\scriptsize
\caption{\CPBXResultCaption{MOCV}{40}{Macro-F1}{coarse}}
\label{tab:cross_mocv_40x_f1_coarse}

\fontsize{7}{8.6}\selectfont
\setlength{\tabcolsep}{1.5pt}
\renewcommand{\arraystretch}{1.0}


\end{table}

\begin{table}[H]
\centering
\setlength{\abovecaptionskip}{4pt}
\setlength{\belowcaptionskip}{2pt}
\scriptsize
\caption{\CPBXResultCaption{MOCV}{40}{Macro-AUROC}{fine}}
\label{tab:cross_mocv_40x_auroc_fine}

\fontsize{7}{8.6}\selectfont
\setlength{\tabcolsep}{1.5pt}
\renewcommand{\arraystretch}{1.0}


\end{table}

\begin{table}[H]
\centering
\setlength{\abovecaptionskip}{4pt}
\setlength{\belowcaptionskip}{2pt}
\scriptsize
\caption{\CPBXResultCaption{MOCV}{40}{Macro-AUROC}{coarse}}
\label{tab:cross_mocv_40x_auroc_coarse}

\fontsize{7}{8.6}\selectfont
\setlength{\tabcolsep}{1.5pt}
\renewcommand{\arraystretch}{1.0}


\end{table}

\begin{table}[H]
\centering
\setlength{\abovecaptionskip}{4pt}
\setlength{\belowcaptionskip}{2pt}
\scriptsize
\caption{\CPBXResultCaption{MOCV}{40}{Macro-AUPRC}{fine}}
\label{tab:cross_mocv_40x_auprc_fine}

\fontsize{7}{8.6}\selectfont
\setlength{\tabcolsep}{1.5pt}
\renewcommand{\arraystretch}{1.0}


\end{table}

\begin{table}[H]
\centering
\setlength{\abovecaptionskip}{4pt}
\setlength{\belowcaptionskip}{2pt}
\scriptsize
\caption{\CPBXResultCaption{MOCV}{40}{Macro-AUPRC}{coarse}}
\label{tab:cross_mocv_40x_auprc_coarse}

\fontsize{7}{8.6}\selectfont
\setlength{\tabcolsep}{1.5pt}
\renewcommand{\arraystretch}{1.0}


\end{table}


\begin{table}[H]
\centering
\setlength{\abovecaptionskip}{4pt}
\setlength{\belowcaptionskip}{2pt}
\scriptsize
\caption{\CPBXResultCaption{MOCV}{20}{Macro-F1}{fine}}
\label{tab:cross_mocv_20x_f1_fine}

\fontsize{7}{8.6}\selectfont
\setlength{\tabcolsep}{1.5pt}
\renewcommand{\arraystretch}{1.0}


\end{table}

\begin{table}[H]
\centering
\setlength{\abovecaptionskip}{4pt}
\setlength{\belowcaptionskip}{2pt}
\scriptsize
\caption{\CPBXResultCaption{MOCV}{20}{Macro-F1}{coarse}}
\label{tab:cross_mocv_20x_f1_coarse}

\fontsize{7}{8.6}\selectfont
\setlength{\tabcolsep}{1.5pt}
\renewcommand{\arraystretch}{1.0}


\end{table}

\begin{table}[H]
\centering
\setlength{\abovecaptionskip}{4pt}
\setlength{\belowcaptionskip}{2pt}
\scriptsize
\caption{\CPBXResultCaption{MOCV}{20}{Macro-AUROC}{fine}}
\label{tab:cross_mocv_20x_auroc_fine}

\fontsize{7}{8.6}\selectfont
\setlength{\tabcolsep}{1.5pt}
\renewcommand{\arraystretch}{1.0}


\end{table}

\begin{table}[H]
\centering
\setlength{\abovecaptionskip}{4pt}
\setlength{\belowcaptionskip}{2pt}
\scriptsize
\caption{\CPBXResultCaption{MOCV}{20}{Macro-AUROC}{coarse}}
\label{tab:cross_mocv_20x_auroc_coarse}

\fontsize{7}{8.6}\selectfont
\setlength{\tabcolsep}{1.5pt}
\renewcommand{\arraystretch}{1.0}


\end{table}

\begin{table}[H]
\centering
\setlength{\abovecaptionskip}{4pt}
\setlength{\belowcaptionskip}{2pt}
\scriptsize
\caption{\CPBXResultCaption{MOCV}{20}{Macro-AUPRC}{fine}}
\label{tab:cross_mocv_20x_auprc_fine}

\fontsize{7}{8.6}\selectfont
\setlength{\tabcolsep}{1.5pt}
\renewcommand{\arraystretch}{1.0}


\end{table}

\begin{table}[H]
\centering
\setlength{\abovecaptionskip}{4pt}
\setlength{\belowcaptionskip}{2pt}
\scriptsize
\caption{\CPBXResultCaption{MOCV}{20}{Macro-AUPRC}{coarse}}
\label{tab:cross_mocv_20x_auprc_coarse}

\fontsize{7}{8.6}\selectfont
\setlength{\tabcolsep}{1.5pt}
\renewcommand{\arraystretch}{1.0}


\end{table}


\begin{table}[H]
\centering
\setlength{\abovecaptionskip}{4pt}
\setlength{\belowcaptionskip}{2pt}
\scriptsize
\caption{\CPBXResultCaption{MOCV}{10}{Macro-F1}{fine}}
\label{tab:cross_mocv_10x_f1_fine}

\fontsize{7}{8.6}\selectfont
\setlength{\tabcolsep}{1.5pt}
\renewcommand{\arraystretch}{1.0}


\end{table}

\begin{table}[H]
\centering
\setlength{\abovecaptionskip}{4pt}
\setlength{\belowcaptionskip}{2pt}
\scriptsize
\caption{\CPBXResultCaption{MOCV}{10}{Macro-F1}{coarse}}
\label{tab:cross_mocv_10x_f1_coarse}

\fontsize{7}{8.6}\selectfont
\setlength{\tabcolsep}{1.5pt}
\renewcommand{\arraystretch}{1.0}


\end{table}

\begin{table}[H]
\centering
\setlength{\abovecaptionskip}{4pt}
\setlength{\belowcaptionskip}{2pt}
\scriptsize
\caption{\CPBXResultCaption{MOCV}{10}{Macro-AUROC}{fine}}
\label{tab:cross_mocv_10x_auroc_fine}

\fontsize{7}{8.6}\selectfont
\setlength{\tabcolsep}{1.5pt}
\renewcommand{\arraystretch}{1.0}


\end{table}

\begin{table}[H]
\centering
\setlength{\abovecaptionskip}{4pt}
\setlength{\belowcaptionskip}{2pt}
\scriptsize
\caption{\CPBXResultCaption{MOCV}{10}{Macro-AUROC}{coarse}}
\label{tab:cross_mocv_10x_auroc_coarse}

\fontsize{7}{8.6}\selectfont
\setlength{\tabcolsep}{1.5pt}
\renewcommand{\arraystretch}{1.0}


\end{table}

\begin{table}[H]
\centering
\setlength{\abovecaptionskip}{4pt}
\setlength{\belowcaptionskip}{2pt}
\scriptsize
\caption{\CPBXResultCaption{MOCV}{10}{Macro-AUPRC}{fine}}
\label{tab:cross_mocv_10x_auprc_fine}

\fontsize{7}{8.6}\selectfont
\setlength{\tabcolsep}{1.5pt}
\renewcommand{\arraystretch}{1.0}


\end{table}

\begin{table}[H]
\centering
\setlength{\abovecaptionskip}{4pt}
\setlength{\belowcaptionskip}{2pt}
\scriptsize
\caption{\CPBXResultCaption{MOCV}{10}{Macro-AUPRC}{coarse}}
\label{tab:cross_mocv_10x_auprc_coarse}

\fontsize{7}{8.6}\selectfont
\setlength{\tabcolsep}{1.5pt}
\renewcommand{\arraystretch}{1.0}


\end{table}



\begin{table}[H]
\centering
\setlength{\abovecaptionskip}{4pt}
\setlength{\belowcaptionskip}{2pt}
\scriptsize
\caption{\CPBXResultCaption{LOOO}{40}{Macro-F1}{fine}}
\label{tab:cross_looo_40x_f1_fine}

\fontsize{7}{8.6}\selectfont
\setlength{\tabcolsep}{1.5pt}
\renewcommand{\arraystretch}{1.0}


\end{table}

\begin{table}[H]
\centering
\setlength{\abovecaptionskip}{4pt}
\setlength{\belowcaptionskip}{2pt}
\scriptsize
\caption{\CPBXResultCaption{LOOO}{40}{Macro-F1}{coarse}}
\label{tab:cross_looo_40x_f1_coarse}

\fontsize{7}{8.6}\selectfont
\setlength{\tabcolsep}{1.5pt}
\renewcommand{\arraystretch}{1.0}


\end{table}

\begin{table}[H]
\centering
\setlength{\abovecaptionskip}{4pt}
\setlength{\belowcaptionskip}{2pt}
\scriptsize
\caption{\CPBXResultCaption{LOOO}{40}{Macro-AUROC}{fine}}
\label{tab:cross_looo_40x_auroc_fine}

\fontsize{7}{8.6}\selectfont
\setlength{\tabcolsep}{1.5pt}
\renewcommand{\arraystretch}{1.0}


\end{table}

\begin{table}[H]
\centering
\setlength{\abovecaptionskip}{4pt}
\setlength{\belowcaptionskip}{2pt}
\scriptsize
\caption{\CPBXResultCaption{LOOO}{40}{Macro-AUROC}{coarse}}
\label{tab:cross_looo_40x_auroc_coarse}

\fontsize{7}{8.6}\selectfont
\setlength{\tabcolsep}{1.5pt}
\renewcommand{\arraystretch}{1.0}


\end{table}

\begin{table}[H]
\centering
\setlength{\abovecaptionskip}{4pt}
\setlength{\belowcaptionskip}{2pt}
\scriptsize
\caption{\CPBXResultCaption{LOOO}{40}{Macro-AUPRC}{fine}}
\label{tab:cross_looo_40x_auprc_fine}

\fontsize{7}{8.6}\selectfont
\setlength{\tabcolsep}{1.5pt}
\renewcommand{\arraystretch}{1.0}


\end{table}

\begin{table}[H]
\centering
\setlength{\abovecaptionskip}{4pt}
\setlength{\belowcaptionskip}{2pt}
\scriptsize
\caption{\CPBXResultCaption{LOOO}{40}{Macro-AUPRC}{coarse}}
\label{tab:cross_looo_40x_auprc_coarse}

\fontsize{7}{8.6}\selectfont
\setlength{\tabcolsep}{1.5pt}
\renewcommand{\arraystretch}{1.0}


\end{table}


\begin{table}[H]
\centering
\setlength{\abovecaptionskip}{4pt}
\setlength{\belowcaptionskip}{2pt}
\scriptsize
\caption{\CPBXResultCaption{LOOO}{20}{Macro-F1}{fine}}
\label{tab:cross_looo_20x_f1_fine}

\fontsize{7}{8.6}\selectfont
\setlength{\tabcolsep}{1.5pt}
\renewcommand{\arraystretch}{1.0}


\end{table}

\begin{table}[H]
\centering
\setlength{\abovecaptionskip}{4pt}
\setlength{\belowcaptionskip}{2pt}
\scriptsize
\caption{\CPBXResultCaption{LOOO}{20}{Macro-F1}{coarse}}
\label{tab:cross_looo_20x_f1_coarse}

\fontsize{7}{8.6}\selectfont
\setlength{\tabcolsep}{1.5pt}
\renewcommand{\arraystretch}{1.0}


\end{table}

\begin{table}[H]
\centering
\setlength{\abovecaptionskip}{4pt}
\setlength{\belowcaptionskip}{2pt}
\scriptsize
\caption{\CPBXResultCaption{LOOO}{20}{Macro-AUROC}{fine}}
\label{tab:cross_looo_20x_auroc_fine}

\fontsize{7}{8.6}\selectfont
\setlength{\tabcolsep}{1.5pt}
\renewcommand{\arraystretch}{1.0}


\end{table}

\begin{table}[H]
\centering
\setlength{\abovecaptionskip}{4pt}
\setlength{\belowcaptionskip}{2pt}
\scriptsize
\caption{\CPBXResultCaption{LOOO}{20}{Macro-AUROC}{coarse}}
\label{tab:cross_looo_20x_auroc_coarse}

\fontsize{7}{8.6}\selectfont
\setlength{\tabcolsep}{1.5pt}
\renewcommand{\arraystretch}{1.0}


\end{table}

\begin{table}[H]
\centering
\setlength{\abovecaptionskip}{4pt}
\setlength{\belowcaptionskip}{2pt}
\scriptsize
\caption{\CPBXResultCaption{LOOO}{20}{Macro-AUPRC}{fine}}
\label{tab:cross_looo_20x_auprc_fine}

\fontsize{7}{8.6}\selectfont
\setlength{\tabcolsep}{1.5pt}
\renewcommand{\arraystretch}{1.0}


\end{table}

\begin{table}[H]
\centering
\setlength{\abovecaptionskip}{4pt}
\setlength{\belowcaptionskip}{2pt}
\scriptsize
\caption{\CPBXResultCaption{LOOO}{20}{Macro-AUPRC}{coarse}}
\label{tab:cross_looo_20x_auprc_coarse}

\fontsize{7}{8.6}\selectfont
\setlength{\tabcolsep}{1.5pt}
\renewcommand{\arraystretch}{1.0}


\end{table}


\begin{table}[H]
\centering
\setlength{\abovecaptionskip}{4pt}
\setlength{\belowcaptionskip}{2pt}
\scriptsize
\caption{\CPBXResultCaption{LOOO}{10}{Macro-F1}{fine}}
\label{tab:cross_looo_10x_f1_fine}

\fontsize{7}{8.6}\selectfont
\setlength{\tabcolsep}{1.5pt}
\renewcommand{\arraystretch}{1.0}


\end{table}

\begin{table}[H]
\centering
\setlength{\abovecaptionskip}{4pt}
\setlength{\belowcaptionskip}{2pt}
\scriptsize
\caption{\CPBXResultCaption{LOOO}{10}{Macro-F1}{coarse}}
\label{tab:cross_looo_10x_f1_coarse}

\fontsize{7}{8.6}\selectfont
\setlength{\tabcolsep}{1.5pt}
\renewcommand{\arraystretch}{1.0}


\end{table}

\begin{table}[H]
\centering
\setlength{\abovecaptionskip}{4pt}
\setlength{\belowcaptionskip}{2pt}
\scriptsize
\caption{\CPBXResultCaption{LOOO}{10}{Macro-AUROC}{fine}}
\label{tab:cross_looo_10x_auroc_fine}

\fontsize{7}{8.6}\selectfont
\setlength{\tabcolsep}{1.5pt}
\renewcommand{\arraystretch}{1.0}


\end{table}

\begin{table}[H]
\centering
\setlength{\abovecaptionskip}{4pt}
\setlength{\belowcaptionskip}{2pt}
\scriptsize
\caption{\CPBXResultCaption{LOOO}{10}{Macro-AUROC}{coarse}}
\label{tab:cross_looo_10x_auroc_coarse}

\fontsize{7}{8.6}\selectfont
\setlength{\tabcolsep}{1.5pt}
\renewcommand{\arraystretch}{1.0}


\end{table}

\begin{table}[H]
\centering
\setlength{\abovecaptionskip}{4pt}
\setlength{\belowcaptionskip}{2pt}
\scriptsize
\caption{\CPBXResultCaption{LOOO}{10}{Macro-AUPRC}{fine}}
\label{tab:cross_looo_10x_auprc_fine}

\fontsize{7}{8.6}\selectfont
\setlength{\tabcolsep}{1.5pt}
\renewcommand{\arraystretch}{1.0}


\end{table}

\begin{table}[H]
\centering
\setlength{\abovecaptionskip}{4pt}
\setlength{\belowcaptionskip}{2pt}
\scriptsize
\caption{\CPBXResultCaption{LOOO}{10}{Macro-AUPRC}{coarse}}
\label{tab:cross_looo_10x_auprc_coarse}

\fontsize{7}{8.6}\selectfont
\setlength{\tabcolsep}{1.5pt}
\renewcommand{\arraystretch}{1.0}


\end{table}

\endgroup

\end{document}